\documentclass{article}

\usepackage{arxiv}

\usepackage[utf8]{inputenc} 
\usepackage[T1]{fontenc}    
\usepackage{hyperref}       
\usepackage{url}            
\usepackage{booktabs}       
\usepackage{amsmath}
\usepackage{amsfonts}       
\usepackage{nicefrac}       
\usepackage{microtype}      
\usepackage{cleveref}       
\usepackage{graphicx}
\usepackage{natbib}
\usepackage{doi}
\usepackage{bm}             
\usepackage[misc]{ifsym}    
\usepackage{caption}
\usepackage{subcaption}
\usepackage{float}          
\usepackage{chngcntr}       
\usepackage{tikz}
\usepackage{pgfplots}
\usepgfplotslibrary{dateplot, groupplots}
\pgfplotsset{compat=newest}
\usepackage{plotpolicies}   

\renewcommand{\cite}[1]{\citep{#1}}

\newcommand{\foi}{\lambda}

\newcommand{\mdpstatespace}{\mathcal{S}}
\newcommand{\mdpactionspace}{\mathcal{A}}
\newcommand{\mdpdiscount}{\gamma}
\newcommand{\mdppolicy}{\pi}
\newcommand{\mdpvaluefunction}{V}
\newcommand{\momdpvaluefunction}{\mathbf{V}}

\newcommand{\mdptransition}{\mathcal{T}}
\newcommand{\mdprewardfn}{\mathcal{R}}
\newcommand{\momdprewardfn}{\bm{\mathcal{R}}}
\newcommand{\state}{s}
\newcommand{\mdpstate}{\mathbf{s}}
\newcommand{\mdpreward}{\mathbf{r}}
\newcommand{\momdpreturn}{\mathbf{R}}
\newcommand{\action}{a}
\newcommand{\mdpaction}{\mathbf{a}}
\newcommand{\chome}{C_{\text{home}}}
\newcommand{\cwork}{C_{\text{work}}}
\newcommand{\ctransport}{C_{\text{transport}}}
\newcommand{\cschool}{C_{\text{school}}}
\newcommand{\cleisure}{C_{\text{leisure}}}
\newcommand{\cother}{C_{\text{other}}}
\newcommand{\momdpname}{MOBelCov}
\newcommand{\agegroups}{K}

\title{Exploring the Pareto front of multi-objective COVID-19 mitigation policies using reinforcement learning}

\date{}

\author{Mathieu Reymond \\
	Vrije Universiteit Brussel\\
	Brussels, Belgium \\
	\texttt{mathieu.reymond@vub.be} \\
	\And
	Conor F. Hayes \\
	National University of Ireland Galway\\
	Galway, Ireland\\
	\And
	Lander Willem \\
	University of Antwerp \\
	Antwerp, Belgium \\
	\And
	Roxana R\u{a}dulescu \\
	Vrije Universiteit Brussel\\
	Brussels, Belgium \\
	\And
	Steven Abrams \\
	Hasselt University \\
	Hasselt, Belgium \\
	\And
	Diederik M.\ Roijers \\
	HU University of Applied Sciences Utrecht \\
	Utrecht, the Netherlands \\
	\And
	Enda Howley \\
	National University of Ireland Galway\\
	Galway, Ireland\\
	\And
	Patrick Mannion \\
	National University of Ireland Galway\\
	Galway, Ireland\\
	\And
	Niel Hens \\
	Hasselt University \\
	Hasselt, Belgium \\
	\And
	Ann Now\'{e} \\
	Vrije Universiteit Brussel\\
	Brussels, Belgium \\
	\And
	Pieter Libin \\
	Vrije Universiteit Brussel\\
	Brussels, Belgium \\
}

\renewcommand{\shorttitle}{Exploring the Pareto front of multi-objective COVID-19 mitigation policies}

\hypersetup{
pdftitle={Exploring the Pareto front of multi-objective COVID-19 mitigation policies using reinforcement learning},
pdfauthor={Mathieu Reymond, Conor F.~Hayes, Lander Willem, Roxana R\u{a}dulescu, Steven Abrams , Diederik M.\ Roijers, Enda Howley, Patrick Mannion, Niel Hens, Ann Now\'{e}, Pieter Libin},
pdfkeywords={Multi-objective reinforcement learning, Epidemic control, COVID-19 epidemic models},
}

\begin{document}
\maketitle

\begin{abstract}
Infectious disease outbreaks can have a disruptive impact on public health and societal processes. As decision making in the context of epidemic mitigation is hard, reinforcement learning provides a methodology to automatically learn prevention strategies in combination with complex epidemic models. Current research focuses on optimizing policies with respect to a single objective, such as the pathogen's attack rate. However, as the mitigation of epidemics involves distinct, and possibly conflicting, criteria (i.a., prevalence, mortality, morbidity, cost), a multi-objective decision approach is warranted to learn balanced policies. To lift this decision-making process to real-world epidemic models, we apply deep multi-objective reinforcement learning and build upon a state-of-the-art algorithm, Pareto Conditioned Networks (PCN), to learn a set of solutions that approximates the Pareto front of the decision problem. We consider the first wave of the Belgian COVID-19 epidemic, which was mitigated by a lockdown, and study different deconfinement strategies, aiming to minimize both COVID-19 cases (i.e., infections and hospitalizations) and the societal burden that is induced by the applied mitigation measures. We contribute a multi-objective Markov decision process that encapsulates the stochastic compartment model that was used to inform policy makers during the COVID-19 epidemic. As these social mitigation measures are implemented in a continuous action space that modulates the contact matrix of the age-structured epidemic model, we extend PCN to this setting. We evaluate the solution set that PCN returns, and observe that it correctly learns to reduce the social burden whenever the hospitalization rates are sufficiently low. In this work, we thus demonstrate that multi-objective reinforcement learning is attainable in complex epidemiological models and provides essential insights to balance complex mitigation policies.
\end{abstract}


\section{Introduction}
As shown by the COVID-19 pandemic, infectious disease outbreaks represent a paramount global challenge that should be tackled by prevention strategies. To this end, understanding the complex dynamics that underlie these epidemics is essential. Epidemiological transmission models allow us to capture and understand such dynamics and facilitate the study of prevention strategies through simulation. However, developing efficient mitigation strategies remains a challenging process, given the non-linear and complex nature of epidemics. 

For this reason, reinforcement learning provides a methodology to automatically learn prevention strategies in combination with complex epidemic models~\cite{libin2020}.
Previous research typically focuses on optimising policies with respect to a single objective, such as the pathogen's attack rate, while the mitigation of epidemics is a problem that inherently covers distinct  and possibly conflicting criteria (i.a., prevalence, mortality, morbidity, cost). Therefore, optimizing on a single objective requires that  these distinct criteria are somehow aggregated into a single metric.
Generally, during learning, a decision maker will rely on an expert's assistance to manually design such a metric. However, a decision maker may
be unaware of all possible trade-offs between different solutions, which may leave the decision maker unaware of possible solutions that better reflect their preferences, potentially resulting in sub-optimal performance. Furthermore, manually designing such metrics is time consuming, costly and error-prone, as this non-intuitive process requires repetitive and tedious tuning to achieve the desired behaviour~\cite{hayes2021practical}. Moreover, taking a single objective approach has several other limiting factors~\cite{hayes2021practical,roijers2013survey}. 

To alleviate the challenging process of defining the learning problem explicitly, we can directly learn optimal behaviours by explicitly taking a multi-objective approach. By assuming that a decision maker will always prefer solutions for which at least one objective improves, it is possible to learn a set of optimal solutions called the \emph{Pareto front}. This enables decision makers to review each solution on the Pareto front before making a decision, thereby being aware of the trade-offs that a solution may imply. 

In this work, we investigate the use of \emph{multi-objective reinforcement learning} (MORL) to learn a set of solutions that approximate the Pareto front of multi-objective mitigation strategies. We consider the first wave of the Belgian COVID-19 epidemic, which was mitigated by a hard lockdown~\cite{willem2021impact}. When the incidence of confirmed cases were steadily dropping, epidemiological experts were asked to investigate strategies to exit from the stringent lockdown which was imposed.
Here, we consider the epidemiological model developed by \citet{abrams2021modelling} that was constructed to describe the Belgian COVID-19 epidemic, and was fitted to hospitalisation incidence data and serial sero-prevalence data.
This model constitutes a stochastic discrete-time age-structured compartmental model that simulates mitigation strategies by varying social distancing parameters concerning school, work and leisure contacts.  
Based on this model, we \emph{contribute \momdpname, a novel mutli-objective epidemiological environment}, in the form of a multi-objective Markov decision process (MOMDP). \momdpname\ encapsulates the epidemiological model developed by \citet{abrams2021modelling} to implement state transitions, with an action space that combines a proportional reduction of school, work and leisure contacts at each time step and defines a reward function based on two objectives: the attack rate (i.e., proportion of the population affected by the pathogen) and social burden.

To learn and explore the trade-offs between the attack rate and social burden we build upon a state-of-the-art MORL approach, namely Pareto Conditioned Networks (PCN)~\cite{reymond2022pcn}, which uses a single neural network to learn the policies that belong to the Pareto front.
As PCN is an algorithm designed for discrete action-spaces, we extend it towards continuous action-spaces to accommodate \momdpname's action-space. With this continuous action variant of PCN, we explore the Pareto front of multi-objective COVID-19 mitigation policies.  

By evaluating the solution set of mitigation policies returned by PCN, we observe that PCN minimises the social burden in scenarios where hospitalization rates are sufficiently low. Therefore, in this work we illustrate that multi-objective reinforcement learning can be utilised to provide important insights surrounding the trade-offs between complex mitigation polices in real-world epidemiological models.

\section{Background}
\subsection{Multi-Objective Reinforcement Learning}
Real-world decisions problems often have multiple and possibly conflicting objectives. Multi-objective reinforcement learning (MORL) can be used to find optimal solutions for sequential decision making problems with multiple objectives~\cite{hayes2021practical}. For MORL, problems are typically modelled as a multi-objective Markov decision process (MOMDP), i.e., a tuple, $\mathcal{M} = (\mdpstatespace, \mdpactionspace, \mdptransition, \mdpdiscount, \momdprewardfn)$, where $\mdpstatespace$, $\mdpactionspace$ are the state and action spaces respectively, $\mdptransition \colon \mdpstatespace \times \mdpactionspace \times \mdpstatespace  \to \left[ 0, 1 \right]$ is a probabilistic transition function, $\mdpdiscount$ is a discount factor determining the importance of future rewards and $\momdprewardfn \colon \mdpstatespace \times \mdpactionspace \times \mdpstatespace \to \mathbb{R}^n$ is an $n$-dimensional vector-valued immediate reward function, where $n$ denotes the number of objectives. For single-objective RL, i.e., when $n = 1$, the goal is to find the policy $\mdppolicy^{*}$ that maximizes the expected sum of discounted rewards, also called the expected return:
\begin{equation}
\mdppolicy^*= \arg\max_\mdppolicy \mathbb{E} \left [ \sum_{t=0}^h \mdpdiscount^t r_t |\ \mdppolicy, s_0 \right ],
\end{equation}
where $r_t = \mdprewardfn(\state_t, \action_t, \state_{t+1})$.
In contrast, in MORL, $n > 1$ which leads to vectorial returns. In this case, there can be policies for which, without any additional information, it is impossible to know if one is better than the other. For example, we cannot decide which policy between $\mdppolicy_1, \mdppolicy_2$ is optimal if they lead to expected returns (also called V-values) $\momdpvaluefunction^{\mdppolicy_1}=(0,1), \momdpvaluefunction^{\mdppolicy_2}=(1,0)$ respectively. We call these solutions \emph{non-dominated}, i.e., solutions for which it is impossible to improve an objective without hampering another. The set that contains all the non-dominated solutions of the decision problem is called the \emph{Pareto front} $\mathcal{F}$. Our goal is to find the set of policies that lead to all the V-values contained in the Pareto front $\Pi^* = \{\ \pi | \momdpvaluefunction^{\pi} \in \mathcal{F}\}$. In general, we call any set of V-values a \emph{solution set}. When a solution set contains only  non-dominated V-values, it is referred to as a \emph{coverage set}. In the case that no $\mdppolicy$ exists that has a $\momdpvaluefunction^{\pi}$ dominating any of the solutions in a coverage set, then this coverage set is the Pareto front.

\subsection{Multi-Objective Metrics}
Comparing the learned coverage sets of different algorithms is a non-trivial task, as one algorithm's output might dominate the other in some region of the objective-space, but be dominated in another. Intuitively, one would generally prefer the algorithm that covers a wider range of decision maker preferences. In this work we use several metrics to evaluate our algorithm's performance.

A widely used metric in the literature is called the \emph{hypervolume}~\cite{zitzler2003}. This metric evaluates the learned coverage set by computing its volume with respect to a fixed specified reference point. This metric is, by definition, the highest for the Pareto front, as no other possible solution can increase its volume (since they are all dominated). One disadvantage of the hypervolume is that it can be difficult to interpret. For example, when working in high-dimensional objective-spaces, adding or removing a single point can lead to wildly different hypervolume values, especially if the point lies close to an extremum of the space.

To alleviate these shortcomings, we additionally evaluate our work on a different metric called the $\varepsilon$-indicator $I_\varepsilon$~\cite{zitzler2003}, which measures how close a coverage set is to the Pareto front $\mathcal{F}$. Intuitively, $I_\varepsilon$ shows that any solution of $\mathcal{F}$ is \emph{at most} $\varepsilon$ better with respect to each objective $o$ than the closest solution of the evaluated coverage set:

\begin{equation}
\label{eq:epsilon-metric}
    I_{\varepsilon} = 
    \inf_{\varepsilon\in\mathbb{R}} 
    \{ 
    \forall\ \momdpvaluefunction^\pi\!{\in}\ \mathcal{F}, ~
    \exists\ \momdpvaluefunction^{\pi'}\!{\in}\ \hat{\Pi} :\ 
    || \mdpvaluefunction^\pi - \mdpvaluefunction_o^{\pi'} ||_{\infty} \le \varepsilon
    \}
\end{equation}

The main disadvantage of this metric is that we need the true Pareto front to compute it, which in the case of our MOMDP (see Section~\ref{sec:sars-cov2-momdp}) is unknown. To still gain insights from our learned policies, we approximate the true Pareto front using the non-dominated policies across all runs.

We note that the $\varepsilon$-indicator metric is quite pessimistic, as it measures worst-case performance~\cite{zintgraf2015}, e.g., it will still report bad performance as long as a single point of the Pareto front is not correctly modeled, even if all the other points are covered. As such, we also use the $I_{\varepsilon-mean}$ \cite{reymond2022pcn} which measures the \emph{average} $\varepsilon$ distance of the solutions in $\mathcal{F}$ with respect to the evaluated coverage set.

\begin{figure}[t!]
    \centering
    \begin{tikzpicture}
      \begin{axis}[legend pos=south west, xmin=-1, xmax=4.5, ymin=-0.5, ymax=4.5,ytick={0,1,2,3,4,5},scale only axis,width=0.4\textwidth]
      \addplot[only marks,mark=x,mark size=3] coordinates {(-0.5,0)};
      \addplot[only marks,mark=*] coordinates {(1,4)(2,2)(4,1)};
      \addplot[only marks,mark=o] coordinates {(0.5,3)(0.75,2.3)(2.3,1)(3.3,0.7)};
      \fill [cyan,opacity=0.2] (-0.5,0) -- (-0.5,3) -- (0.5,3) -- (0.5,2.3) -- (0.75,2.3) --
      (0.75,1) -- (2.3,1) -- (2.3,0.7) -- (3.3,0.7) -- (3.3,0) -- cycle;
      \draw [dashed, -|] (0.5, 3) -- node[right]{$\varepsilon_1$} (0.5,4);
      \draw [dashed, -|] (2.3,1) -- node[right]{$\varepsilon_2$} (2.3,2);
      \draw [dashed, -|] (3.3,0.7) -- node[below]{$\varepsilon_3$} (4,0.7);
      \draw [opacity=0.2,line width=2,orange, dotted] (-0.5,0) -- (-0.5,4) -- (1,4) -- (1,2) --
      (2,2) -- (2,1) -- (4,1) -- (4,0) -- (-0.5,0);
      \end{axis}
    \end{tikzpicture}
    \caption{Example of a Pareto front (black dots) and a coverage set (white dots) in a 2-objective environment. The hypervolume metric (in light blue) measures the volume of all dominated solutions with respect to some reference point (cross). The $\varepsilon$ metrics first compute the maximum distance between each point in the Pareto front and its closest point in the coverage set ($\varepsilon_i$). We can then take their maximum value to compute the $I_\varepsilon$ metric, or their mean value to obtain the $I_{\varepsilon-mean}$ metric of the coverage set.}
    \label{fig:paretofrontmetrics}
\end{figure}
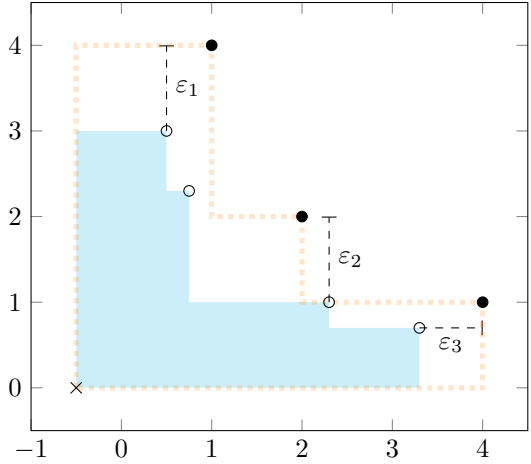

Figure~\ref{fig:paretofrontmetrics} shows a visual representation of the hypervolume and $\varepsilon$ metrics in two dimensions.

\section{COVID-19 model and the \momdpname\  MOMDP}
\label{sec:sars-cov2-momdp}
\subsection{Stochastic compartment model for SARS-CoV-2}
\label{sec:sars-cov2-momdp_model}
As an environment to evaluate non-pharmaceutical interventions, we consider the adapted version of the SEIR compartmental model presented by Abrams et al., that was used to investigate exit strategies in Belgium after the first epidemic wave of SARS-CoV-2~\cite{abrams2021modelling}. This model constitutes a discrete-time stochastic model, that considers an age-structured population. This model generalises a standard SEIR model\footnote{A standard SEIR model divides the population into four different states, i.e., susceptible, exposed, infectious and recovered individuals.}, extended to capture the different stages of disease spread and history that are associated with SARS-CoV-2 (i.e., pre-symptomatic, asymptomatic, symptomatic with mild symptoms and symptomatic with severe symptoms) and to represent the stages associated with severe disease, i.e., hospitalisation, admission to the intensive care unit (ICU) and death. 
As we consider an age-structured population, we consider this extended SEIR structure for $\agegroups = 10$ age groups,  i.e., $[0-10), [10-20), [20-30), [30-40), [40-50), [50-60), [60-70), [70-80), [80-90), [90,\infty)$. Contacts of the different age-groups which impact the propagation rate of the epidemic are modeled using social contact matrices (SCMs). We define a SCM for 6 different social environments: $\chome, \cwork, \ctransport, \cschool, \cleisure, \cother$ for the home, work, transport, school, leisure, other environments respectively.
The model is described by a set of ordinary differential equations (ODEs), as described in SI.
By formulating this set of differential equations defined above as a chain-binomial process we can obtain stochastic trajectories from this model~\cite{abrams2021modelling}.

\subsection{Interventions strategies}
\label{sec:interventions}
In order to model different types of interventions, we follow \citet{abrams2021modelling}. Firstly, in order to consider distinct exit scenarios, we alter the SCMs to reflect a contact reduction in a particular age group. Secondly, we assume that compliance to the interventions is gradual and model this using a logistic compliance function (see details in SI).
We consider a contact reduction function that imposes a proportional reduction of work (including transport) $p_w$, school $p_s$ and leisure $p_l$ contacts, which is implemented as a linear combination of contact matrices:
\begin{equation}
\label{eq:scm}
    \hat{C} = \chome + p_w (\cwork + \ctransport) + p_s \cschool + p_l (\cleisure + \cother)
\end{equation}

\subsection{The \momdpname\ Environment}
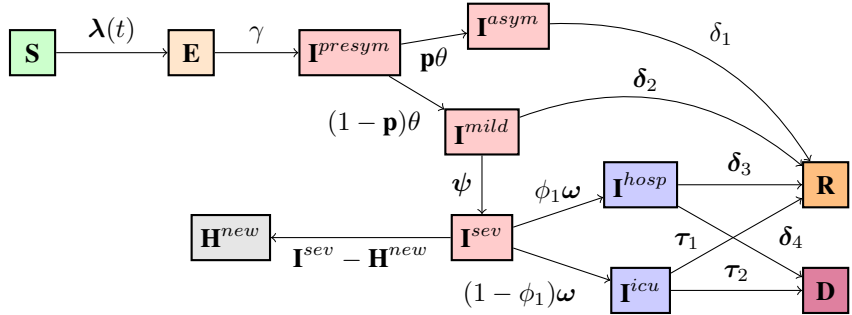
\begin{figure}[t!]
    \centering
    \begin{tikzpicture}
     \draw (0.0pt, 0.0pt)node[fill=green!20, thick, minimum height=0.6cm,minimum width=0.6cm, draw](0){$\textbf{S}$};
     \draw (60.0pt, 0.0pt)node[fill=orange!20, thick, minimum height=0.6cm,minimum width=0.6cm, draw](1){$\textbf{E}$};
     \draw (120.0pt, 0.0pt)node[fill=red!20, thick, minimum height=0.6cm,minimum width=0.6cm, draw](2){$\textbf{I}^{presym}$};
     \draw (180.0pt, 10.0pt)node[fill=red!20, thick, minimum height=0.6cm,minimum width=0.6cm, draw](3){$\textbf{I}^{asym}$};
     \draw (170.0pt, -30.0pt)node[fill=red!20, thick, minimum height=0.6cm,minimum width=0.6cm, draw](4){$\textbf{I}^{mild}$};
     \draw (170.0pt, -70.0pt)node[fill=red!20, thick, minimum height=0.6cm,minimum width=0.6cm, draw](5){$\textbf{I}^{sev}$};
     \draw (230.0pt, -50.0pt)node[fill=blue!20, thick, minimum height=0.6cm,minimum width=0.6cm, draw](6){$\textbf{I}^{hosp}$};
      \draw (75.0pt, -70.0pt)node[fill=gray!20, thick, minimum height=0.6cm,minimum width=0.6cm, draw](10){$\textbf{H}^{new}$};
     \draw (230.0pt, -90.0pt)node[fill=blue!20, thick, minimum height=0.6cm,minimum width=0.6cm, draw](7){$\textbf{I}^{icu}$};
     \draw (300.0pt, -50.0pt)node[fill=orange!50, thick, minimum height=0.6cm,minimum width=0.6cm, draw](8){$\textbf{R}$};
     \draw (300.0pt, -90.0pt)node[fill=purple!50, thick, minimum height=0.6cm,minimum width=0.6cm, draw](9){$\textbf{D}$};
     
     \path[->] (0) edge[] node[above]{$\boldsymbol{\lambda}(t)$} (1);
     \path[->] (1) edge[] node[above]{$\gamma$} (2);
     \path[->] (2) edge[] node[below]{$\textbf{p}\theta$} (3);
     \path[->] (2) edge[] node[near end, below left]{$(1-\textbf{p})\theta$} (4);
     \path[->] (4) edge[] node[left]{$\boldsymbol{\psi}$} (5);
     \path[->] (5) edge[] node[near end, below left]{$(1-\phi_{1})\boldsymbol{\omega}$} (7);
     \path[->] (5) edge[] node[above]{$\phi_{1}\boldsymbol{\omega}$} (6);
     \path[->] (7) edge[] node[near end, below right]{$\boldsymbol{\delta}_{4}$} (8);
     \path[->] (7) edge[] node[above]{$\boldsymbol{\tau}_{2}$} (9);
     \path[->] (6) edge[] node[above]{$\boldsymbol{\delta}_{3}$} (8);
     \path[->] (6) edge[] node[near start, below left]{$\boldsymbol{\tau}_{1}$} (9);
     \path[->] (3) edge[bend left = 30] node[above right]{$\delta_{1}$} (8);
     \path[->] (4) edge[bend left = 30] node[above left]{$\boldsymbol{\delta}_{2}$} (8);
      \path[->] (5) edge[] node[below]{$\textbf{I}^{sev} - \textbf{H}^{new}$} (10);
    \end{tikzpicture}
    
    \caption{Schematic diagram of the compartmental model for SARS-CoV-2 presented by \citet{abrams2021modelling} which is used to derive the MOMDP.}
    \label{fig:ode_mdel}
\end{figure}
In order to apply multi-objective reinforcement learning, we construct the \momdpname\ MOMDP based on the epidemiological model introduced in Section~\ref{sec:sars-cov2-momdp_model} and graphically depicted in Figure~\ref{fig:ode_mdel}.

\paragraph{State-space:} The state of the MOMDP consists of the state variables that make up the epidemiological model (see Figure~\ref{fig:ode_mdel}), to which we refer as $\mdpstate_m$, together with the social contact matrix $\hat{C}$ that is currently in place.
$\mdpstate_m$ directly corresponds to the aggregation of the state variables in the epidemiological model, i.e., a tuple,
\begin{equation}
    \{S_k, E_k, I_k^{presym}, I_k^{asym}, I_k^{mild}, I_k^{sev}, I_k^{hosp}, I_k^{icu}, H^{new}_k, D_k, R_k\},
\end{equation}
for each age group $k \in \{1, \ldots, \agegroups\}$, where $S$ encodes the members of the population who are susceptible to infection and $E$ encodes the members of the population who have been exposed to COVID-19. Moreover, $I^{presym}$, $I^{asym}$, $I^{mild}$, $I^{hosp}$, $I^{icu}$ are the members of the population infected with COVID-19 and are, respectively, presymptomatic, asymptomatic, have mild symptoms, are hospitalised, or are in the ICU. Finally, $H^{new}_k$ represents the number of newly hospitalised individuals in age group $k$.

Therefore, we define a state in \momdpname\ as follows:
\begin{equation}
    \mdpstate = \mdpstate_m \cup \hat{C}
\end{equation}

We paramaterise the model using the mean of the posteriors as reported by \citet{abrams2021modelling} (details in SI).

\paragraph{Action-space:} Our actions concern the installment of a social contact matrix $\hat{C}$, with a particular reduction (see Section~\ref{sec:interventions}). To this end, we use the proportional reduction parameters $p_w, p_s, p_l$ defined in Section~\ref{sec:interventions}. Thus, each $\mdpaction \in \mdpactionspace$ is a 3-dimensional continuous vector in $[0,1]^3$ (i.e., $\mdpaction = [p_{w}, p_{s}, p_{l}]$) which impacts the SCM according to Equation~\ref{eq:scm}.

\paragraph{Transition function:} The model defined by \citet{abrams2021modelling} utilises a model transition probability $M(\mdpstate_m' \mid \mdpstate_m, \hat{C})$ (see SI for details on $M$), where $\hat{C}$ the currently installed SCM, that progresses the epidemiological model in one timestep.
We use this function as the transition function in \momdpname.
For each timestep $t$, we simulate the model for one week, using $\hat{C}$ obtained from $\mdpaction_t$. 

As mentioned above and detailed in the SI, the compartment model is described by a set of ODEs. To obtain a stochastic version, the ODEs can be formulated as a chain-binomial process. Based on these cases, we also create and experiment on two versions of the \momdpname\ environment: the (deterministic) \emph{ODE model} (i.e., a MOMDP with a deterministic transition function) and the (stochastic) \emph{Binomial model} (i.e., a MOMDP with a stochastic transition function).

\paragraph{Reward function:} We define a vector reward function which considers multiple objectives: attack rate (i.e., infections, hospitalisations) and the burden imposed by the interventions on the population.

The attack rate in terms of infections is defined as the difference in susceptibles from the current state and next state~\cite{libin2020}. Since this is a cost that needs to be minimized, we defined the corresponding reward function as the negative attack rate:
\begin{equation}
\mdprewardfn_{\text{ARI}}(\mdpstate,\mdpaction,\mdpstate') = -(\sum_{k=1}^{\agegroups}S_k(\mdpstate)-\sum_{k=1}^{\agegroups}S_k(\mdpstate')).
\end{equation}

The reward function to reduce the attack rate in terms of hospitalisations is simply defined as the negative number of new hospitalizations:
\begin{equation}
\mdprewardfn_{\text{ARH}}(\mdpstate,\mdpaction,\mdpstate') = -\sum_{k=1}^{\agegroups}H^\text{new}_k(\mdpstate)
\label{eqn:attack-rate-hosp}
\end{equation}

Finally, we use the missed contacts resulting from the intervention measures as a proxy for societal burden. To quantify missed contacts, we consider the original social contact matrix $C$ and the installed social contact matrix $\hat{C}$, and compute the difference $\hat{C} - C$. 
The resulting difference matrix quantifies the average frequency of contacts missed. To determine missed contacts for the entire population, we apply the calculated difference matrix to the population sizes of the respective age groups that are currently uninfected (i.e., susceptible and recovered individuals). Therefore we define the social burden reward function $\mdprewardfn_{\text{SB}}$, as follows:

\begin{equation}
    \mdprewardfn_{\text{SB}}(\mdpstate,\mdpaction,\mdpstate') = \sum_{i=1}^{\agegroups}\sum_{j=1}^{\agegroups}(\hat{C}-C)_{ij}S_i(\mdpstate)S_j(\mdpstate) + \sum_{i=1}^{\agegroups}\sum_{j=1}^{\agegroups}(\hat{C}-C)_{ij}R_i(\mdpstate)R_j(\mdpstate),
\end{equation}
where $S_k(\mdpstate)$ represents the number of susceptible individuals in age group $k$ in state $\mdpstate$ and $R_k$ represents the number of recovered individuals in age group $k$ in state $\mdpstate$. In Section~\ref{sec:experiments}, we optimize PCN on two different variants for the multi-objective reward function: $[\mdprewardfn_\text{ARH}, \mdprewardfn_\text{SB}]$ and $[\mdprewardfn_\text{ARI}, \mdprewardfn_\text{SB}]$.


\section{Pareto Conditioned Networks}
In multi-objective optimization, the set of optimal policies can grow exponentially in the number of objectives. Thus, recovering them all is an expensive process and requires an exhaustive exploration of the complete state space. To address this problem, we use Pareto Conditioned Networks~(PCN), a method that uses a single neural network to encompass all non-dominated
policies \cite{reymond2022pcn}. The key idea behind PCN is to use supervised learning techniques to improve the policy instead of resorting to temporal-difference learning. This eliminates the moving-target problem \cite{schmidhuber2019}, resulting in stable learning algorithm.

PCN uses a single neural network that takes a tuple $\langle \mdpstate, \hat{h}, \mathbf{\hat{R}}  \rangle$ as input. $\mathbf{\hat{R}}$ represents the \emph{desired return} of the decision maker, i.e. the return PCN should reach at the end of the episode. $\hat{h}$ denotes the \emph{desired horizon} that expresses the number of timesteps that should be executed before reaching $\mathbf{\hat{R}}$. At execution time, both $\hat{h}$ and $\mathbf{\hat{R}}$ are chosen by the decision maker at the start of the episode. Then, at every timestep, the desired horizon is updated according to the perceived reward $\mdpreward_t$, $\mathbf{\hat{R}} \leftarrow \mathbf{\hat{R}} - \mdpreward_t$ and the desired horizon is decreased by one, $\hat{h} \leftarrow \hat{h}-1$.

PCN's neural network has an output for each action $\action_i \in \mathcal{A}$. Each output represents the confidence the network has that, by taking the corresponding action in $\mdpstate$, $\mathbf{\hat{R}}$ will be reached in $\hat{h}$ timesteps. We can draw an analogy with a classification problem where the network should learn to classify $(\mdpstate, \hat{h}, \mathbf{\hat{R}})$ to its corresponding label $\action_i$.

Similar to classification, PCN requires a labeled dataset with training examples to learn a mapping from input to label. However, contrary to classification, the data present in the dataset is not fixed. Each time PCN executes a trajectory it creates a tuple $x=(\mdpstate_t, \momdpreturn_t, T-t)$, $y=\action_t$ for every transition of the trajectory and adds it to the dataset.

PCN collects new data for a fixed number of episodes, after which it re-trains the network with batch updates from the newly assembled dataset. This improves the policies induced by the network, which in turn enables the agent to gather superior data for the next training batch.

\subsection{PCN for continuous actions}
\label{sec:pcn-continuous}

PCN trains the network as a classification problem, where each class represents a different action. Transitions $x= \langle \mdpstate_t, h_t, \momdpreturn_t \rangle, y=\action_t$ are sampled from the dataset, and the ground-truth output $y$ is compared with the predicted output $\hat{y}=\pi(\mdpstate_t, h_t, \momdpreturn_t)$. The predictor (i.e., the policy) is then updated using the cross-entropy loss function:
\begin{equation}
    H = -\sum_{\action \in \mathcal{A}}{y_{\action} \log \pi(\action|\mdpstate_t,h_t,\momdpreturn_t)}
\end{equation}
where $y_\action = 1$ if $\action = \action_t$ and $y_\action = 0$ otherwise.

While the original PCN algorithm is designed for MOMDPs with discrete actions-spaces, the problem we tackle (see Section~\ref{sec:sars-cov2-momdp}) is defined in terms of a continuous action-space. We thus extend PCN for the continuous action-space setting.
First, we change the output of the neural network such that there is a single output value for each dimension of the action-space. Since the actions should be bound in the domain of possible actions ($[0,1]$ in the case of \momdpname, see Section~\ref{sec:sars-cov2-momdp}), we apply a tanh non-linearity function on this output.
Second, the problem becomes a regression problem instead of a classification problem, as the labeled dataset uses continuous labels $y = \mdpaction_t$ instead of categories. We thus use a Mean Squared Error (MSE) loss to update the policy:

\begin{equation}
    MSE = \frac{1}{|\mathcal{A}|}\sum_{a \in \mathcal{A}}{(\hat{y}_a - y_a)^2}
\end{equation}

Since learning the full set of Pareto-efficient policies $\Pi^*$ requires that the policies $\pi^* \in \Pi^*$ are deterministic stationary policies \cite{roijers2013survey}, we use the output $\mathbf{\hat{y}}$ as action at execution time. However, PCN improves its policy through exploration, by continuously updating its dataset with better trajectories. Thus, at training time, we follow \cite{lillicrap2015continuous} and use a stochastic policy by adding random noise sampled from a Normal distribution to the action: 
\begin{equation}
    \mdpaction_t = \pi(\mdpstate_t, h_t, \momdpreturn_t) + \eta s \text { with } s \sim \mathcal{N},
\end{equation}

where $\eta$ is a hyper-parameter defining the magnitude of noise to be added.

\subsection{Coping with stochastic transitions}

PCN trains its policy on a dataset that is collected by executing trajectories. It assumes that reenacting a transition from the dataset leads to the same episodic return. When the transition function $\mathcal{T}$ of the MOMDP is deterministic, the whole trajectory can be faithfully reenacted, which guarantees that we obtain the same return. Combined with the fact that PCN's policy is deterministic at execution time, conditioning the policy on a target episodic return is equivalent to conditioning it on the V-value $\mathbf{V}$.

However, when $\mdptransition$ is stochastic we lose this guarantee. We cope with this issue by adding small random noise to $\mathbb{R}_t$ when performing gradient descent, which reduces the risk of overfitting~\cite{zur2009noise}. Moreover, although the Binomial model of \momdpname\ is stochastic, the possible next-states resulting from a state-action pair are similar to each other. This allows PCN to compensate if $\mdpreward_t = \momdprewardfn(\mdpstate_t, \mdpaction_t, \mdpstate_{t+1})$ is somewhat worse than expected. This is confirmed by our experiments (see Section~\ref{sec:experiments}), where the coverage sets learned by PCN on the ODE model and on the Binomial model are similar.

\section{Analysing COVID-19 deconfinement policies}
\label{sec:experiments}
Our goal is to use PCN to learn deconfinment strategies in the \momdpname\ environment. We aim to obtain policies that balance between the social burden experienced by the population and the epidemiological objective of minimising the attack rate. To this end we consider two cases for the vectorial reward functions $[\mdprewardfn_\text{ARH}, \mdprewardfn_\text{SB}]$ and $[\mdprewardfn_\text{ARI}, \mdprewardfn_\text{SB}]$, to learn and analyse policies under different targets with respect to the considered attack rate.

We now evaluate our extension of PCN for continuous action-spaces on both the ODE and Binomial models. Conform to \citet{abrams2021modelling}, the simulation starts on the 1st of March 2020, by seeding a number of infections in the population. Two weeks later, on the 14th of March, the Belgian government initiated a full lockdown of the country. This is implemented by fixing the actions $p_w, p_s, p_l$ to $0.2, 0, 0.1$ respectively. This lockdown ended on the 4th of May 2020, at which point the government decided on a multi-phase exit strategy to incrementally reduce teleworking, reopen schools and allow leisure activities, such as the re-opening of bars and the cultural sector. It is from this day onward that PCN aims to learn policies for diverse exit strategies, compromising between the total number of daily new hospitalizations and the total number of contacts lost as a proxy for social burden. The simulation lasts throughout the school holidays (from 01/07/2020 to 31/08/2020). Schools are closed during the school holidays, which is simulated by setting $p_s = 0$, regardless of the corresponding value outputted by the policy, i.e., during periods of school closure $p_s$ is ignored.

To evaluate the quality of the policies discovered by PCN, we compare it to a baseline. The baseline consists of a set of 100 fixed policies, that iterate over all the possible social restriction levels, with values ranging from 0 to 1. In other words, the fixed policies directly operate in a fine-grained manner on the whole contact reduction function $\hat{C}$. This allows us to obtain a strong baseline for potential exit strategies over the objective space. We note that while such fixed policies are a feasible approach, they do not scale well in terms of action and objective spaces and they will not be able to provide an adaptive restriction level, which is what we aim to provide using PCN.

All experiments are averaged over 5 runs. The hyper-parameters and the neural network architecture can be found in the SI.

\begin{figure}
    \centering
    \begin{tikzpicture}
        \pgfplotsset{
          scale only axis,
          height=6cm,
          xlabel={Cumulative number of daily new hospitalizations},
          ylabel={Cumulative lost contacts}
        }
      \begin{axis}[
        legend entries={{$\text{PCN}_\text{ARH}$ (ODE)}, {$\text{PCN}_\text{ARI}$ (ODE)}, {Fixed}},
        width=9.5cm
      ]
      \addplot[only marks,draw opacity=0.5,mark=x,red] table [x=o_1, y=o_5, col sep=comma] {data/cs/arh/cs_ode.csv};
      \addplot[only marks,draw opacity=0.5,mark=o,blue] table [x=o_1, y=o_5, col sep=comma] {data/cs/ari/cs_ode.csv};
      \addplot[only marks,mark=+] table [x=o_0, y=o_1, col sep=comma] {data/cs/cs_fixed.csv};
      \end{axis}
    \end{tikzpicture}
    \caption{The Pareto front of policies discovered by PCN using the ODE model (Binomial omitted for clarity, see SI), showing the different compromises between the number of hospitalizations and the number of lost contacts. Although $\text{PCN}_\text{ARI}$ was trained on the number of infections, it still shows a competitive coverage set with respect to hospitalizations.}
    \label{fig:pareto-front}
\end{figure}
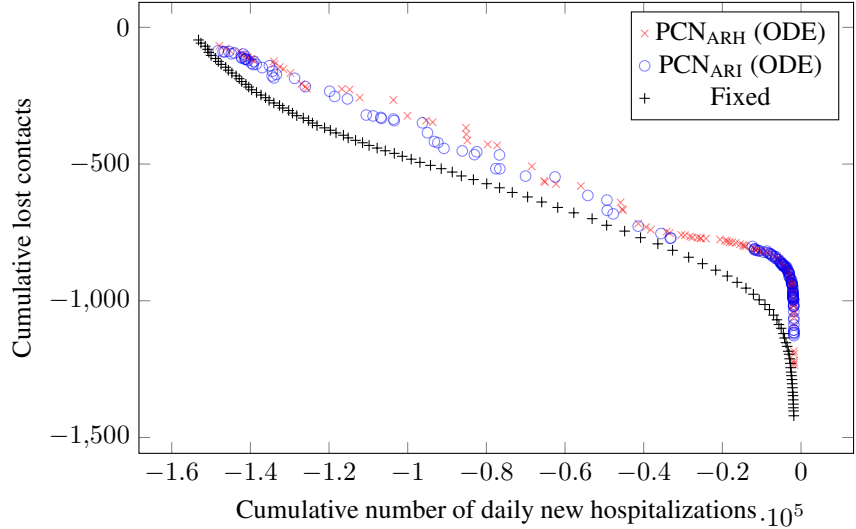

\subsection{Learned coverage set}

\begin{figure}
  \centering
  \begin{tikzpicture}
\pgfplotsset{
      scale only axis,
      date coordinates in=x,
      xticklabel=\day/\month,
      height=2cm,
    }
    \begin{groupplot}[
    group style={
            group size=1 by 3,
            y descriptions at=edge left,
            x descriptions at=edge bottom,
            horizontal sep=0pt,
            vertical sep=0pt
        },
      width=9.5cm,
      axis y line*=left,
      xlabel={time},
      ylabel={people},
    ]
    \nextgroupplot[scaled y ticks=false]
    \addplot[mark=none, cyan] table [x=dates, y=i_hosp_new, col sep=comma] {data/ode/policy_0.csv};
    \label{i_hosp_new}
    \addplot[mark=none, orange] table [x=dates, y=i_icu_new, col sep=comma] {data/ode/policy_0.csv};
    \label{i_icu_new}
    \addplot[mark=none, red] table [x=dates, y=d_new, col sep=comma] {data/ode/policy_0.csv};
    \label{d_new}
    \nextgroupplot[scaled y ticks=false]
    \addplot[mark=none, cyan] table [x=dates, y=i_hosp_new, col sep=comma] {data/ode/policy_30.csv};
    \addplot[mark=none, orange] table [x=dates, y=i_icu_new, col sep=comma] {data/ode/policy_30.csv};
    \addplot[mark=none, red] table [x=dates, y=d_new, col sep=comma] {data/ode/policy_30.csv};
    \nextgroupplot[scaled y ticks=false]
    \addplot[mark=none, cyan] table [x=dates, y=i_hosp_new, col sep=comma] {data/ode/policy_139.csv};
    \addplot[mark=none, orange] table [x=dates, y=i_icu_new, col sep=comma] {data/ode/policy_139.csv};
    \addplot[mark=none, red] table [x=dates, y=d_new, col sep=comma] {data/ode/policy_139.csv};
    \end{groupplot}
    \begin{groupplot}[
    group style={
            group size=1 by 3,
            y descriptions at=edge right,
            x descriptions at=edge bottom,
            horizontal sep=0pt,
            vertical sep=0pt
        },
      width=9.5cm,
      axis y line*=right,
      axis x line=none,
      ymax=1.1,
      ylabel={proportion},
      legend style={at={(0.5,1)},anchor=south},
      legend columns=3, 
      legend cell align=left,
    ]
    \nextgroupplot[scaled y ticks=false]
    \addplot[mark=none, dashed, blue] table [x=dates, y=p_w, col sep=comma] {data/ode/policy_0.csv};
    \label{p_w}
    \addplot[mark=none, dashed, magenta] table [x=dates, y=p_s, col sep=comma] {data/ode/policy_0.csv};
    \label{p_s}
    \addplot[mark=none, dashed, darkgray] table [x=dates, y=p_l, col sep=comma] {data/ode/policy_0.csv};
    \label{p_l}
    \addlegendimage{/pgfplots/refstyle=i_hosp_new}\addlegendentry{$p_w$}
    \addlegendimage{/pgfplots/refstyle=i_icu_new}\addlegendentry{$p_l$}
    \addlegendimage{/pgfplots/refstyle=d_new}\addlegendentry{$p_s$}
    \addlegendimage{/pgfplots/refstyle=p_w}\addlegendentry{{daily new hosp}}
    \addlegendimage{/pgfplots/refstyle=p_s}\addlegendentry{{daily new ICU}}
    \addlegendimage{/pgfplots/refstyle=p_l}\addlegendentry{{daily deaths}}
    \nextgroupplot[scaled y ticks=false]
    \addplot[mark=none, dashed, blue] table [x=dates, y=p_w, col sep=comma] {data/ode/policy_30.csv};
    \addplot[mark=none, dashed, magenta] table [x=dates, y=p_s, col sep=comma] {data/ode/policy_30.csv};
    \addplot[mark=none, dashed, darkgray] table [x=dates, y=p_l, col sep=comma] {data/ode/policy_30.csv};
    \nextgroupplot[scaled y ticks=false]
    \addplot[mark=none, dashed, blue] table [x=dates, y=p_w, col sep=comma] {data/ode/policy_139.csv};
    \addplot[mark=none, dashed, magenta] table [x=dates, y=p_s, col sep=comma] {data/ode/policy_139.csv};
    \addplot[mark=none, dashed, darkgray] table [x=dates, y=p_l, col sep=comma] {data/ode/policy_139.csv};
    \end{groupplot}
  \end{tikzpicture}
  \caption{Selection of policies learned by PCN, from most restrictive in terms of social burden (top) to least restrictive (bottom).
  }
  \label{fig:policy-execution}
\end{figure}
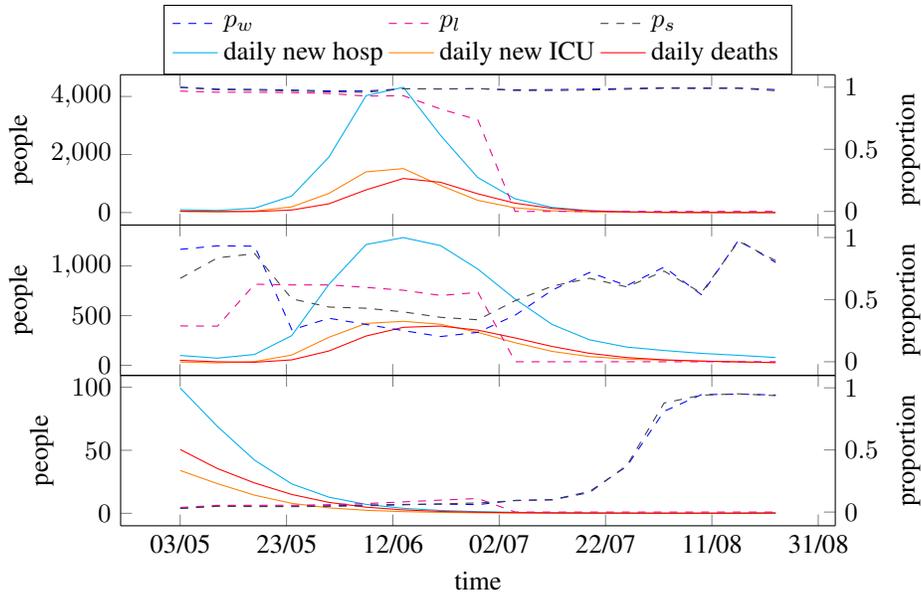

We learn a coverage set that ranges from imposing minimal restrictions to enforcing many restrictions (see Figure~\ref{fig:policy-execution}). The coverage set is shown in Figure~\ref{fig:pareto-front}.

We notice that the coverage sets discovered by PCN almost completely dominate the coverage set of the baseline, showing that there are much better alternatives to the fixed policies. This is most visible in the compromising policies, where one has to carefully choose when to remove social restrictions while at the same time minimizing the impact on daily new hospitalizations. In these scenarios, PCN discovers policies that can drastically reduce the total number of new hospitalizations (e.g. more than $20000$) for the same social burden. Analysing the executions of such policies (see Figure~\ref{fig:policy-execution}, middle plot) shows a flattened hospitalization curve, with a gradual increase of social freedom during the school holidays such that the peak stays stable and gradually decreases over time.

Interestingly, we notice that the most restrictive policy (i.e. the one that prioritizes hospitalizations over social burden, see Figure~\ref{fig:policy-execution}, bottom plot) still starts to gradually increase $p_w$ and $p_l$ from the end of July onward. This is because by then, the epidemic has mostly faded out, and it is safe to reduce social restrictions. The timing of this reduction is important as reducing restrictions too soon can lead to a new wave. PCN learns the impact of its decisions over time, and correctly infers the timing at which restrictions can be safely lifted.

\subsection{ODE versus Binomial models}
\label{sec:ode-vs-binomial}

Next, we compare the performance of PCN when trained on the ODE and Binomial models. The ODE model has the advantage of using a deterministic transition function, for which PCN is suited. However, taking into account stochasticity, using the Binomial model, is important as individual behavioural changes introduce substantial uncertainty in the course of the outbreak and require stochastic model evaluations to evaluate the impact of this uncertainty on the effectiveness of the intervention strategies. Regardless, the results displayed in Table~\ref{tab:coverage-set} show that the solution sets that PCN finds using the Binomial model are competitive with the ODE model, as indicated by their hypervolumes and $I_{\varepsilon-mean}$ metrics. We do note that the difference in $I_{\varepsilon}$ indicator is more significant, indicating that a few solutions discovered using the ODE model are not found when using the Binomial model, impacting worst-case performance. Overall, we see that our extension of PCN performs efficiently, even with some degree of stochasticity in the transition function.

\begin{table}[t]
  \centering
  \setlength{\tabcolsep}{0.5em} 
  {\renewcommand{\arraystretch}{1.2}
  \begin{tabular}{|l|c|c|c|}
    \hline
        & Hypervolume & $I_\varepsilon$ & $I_{\varepsilon-mean}$ \\
    \hline
        $\text{PCN}_\text{ARH}$ (ODE) & $\mathbf{0.679 \pm 0.010}$ & $\mathbf{0.056 \pm 0.004}$ & $\mathbf{0.010 \pm 0.006}$ \\
        \hline
        $\text{PCN}_\text{ARH}$ (Binomial) & $0.674 \pm 0.005$ & $0.085 \pm 0.044$ & $0.017 \pm 0.005$ \\
        \hline
        $\text{PCN}_\text{ARI}$ (ODE) & $0.668 \pm 0.009$ & $0.084 \pm 0.052$ & $0.021 \pm 0.011$ \\
        \hline
        $\text{PCN}_\text{ARI}$ (Binomial) & $0.659 \pm 0.007$ & $0.100 \pm 0.054$ & $0.028 \pm 0.013$ \\
        \hline
        Fixed & $0.6 \pm 0.0$ & $0.106 \pm 0.0$ & $0.058 \pm 0.0$ \\
        \hline
\end{tabular}

  }
  \caption{Evaluation metrics for the coverage sets comparing hospitalizations with social burden. In general training on the ODE results in slightly better coverage sets than on the Binomial model. Training on infections (ARI) still provides a competitive coverage set in terms of hospitalizations. All PCN coverage sets outperform the baseline.}
  \label{tab:coverage-set}
\end{table}

\subsection{$R_\text{ARH}$ versus on $R_\text{ARI}$}
\label{sec:arh-vs-ari}

We now assess the difference in coverage sets when optimizing on $\mdprewardfn_\text{ARH}$ versus $\mdprewardfn_\text{ARI}$. Although at a different scale, our experiments show that infections and hospitalizations are correlated. This is confirmed in Figure~\ref{fig:pareto-front}: the coverage set \emph{in terms of hopsitalizations} learned by $\text{PCN}_\text{ARI}$ is only slightly worse than the one learned by $\text{PCN}_\text{ARH}$, even though $\text{PCN}_\text{ARI}$ optimized \emph{on the infection attack rate}. This is expected, as during the initial phase of the epidemic, limited immunity was present in the population (few natural immunity and no vaccines), which induces a tight coupling between infection and hospitalisation cases. Moreover, the opposite also holds: both coverage sets are similar in terms of \emph{infections}. However, there is one notable exception. A number of policies learned by $\text{PCN}_\text{ARH}$ start removing social restrictions before the epidemic is rooted out, resulting in a rise of hospitalizations at the end of the summer holidays. Since, following Abrams et al., this is when the simulation stops, the total number of hospitalizations remains low. The number of infections, however, starts growing before this rise in hospitalizations and, since its growth is exponential, ends up high. We note that $\text{PCN}_\text{ARH}$ also learns alternative policies for similar $[\mdprewardfn_\text{ARH}, \mdprewardfn_\text{SB}]$ that do not exhibit this behavior. This shows that, even though a policy might lead to a return that matches the preferences of the decision maker, its execution might still lead to undesired behavior. We argue that the use of such expert systems for decision making should be paired with careful interpretation.

\subsection{Robustness of policy executions}
\label{sec:pcn-robustness}

The dataset of trajectories that PCN is trained on is pruned over time to keep only the most relevant trajectories. The returns of these trajectories are used in Figure~\ref{fig:pareto-front} to show the discovered coverage set. Each of these returns can be used as desired return for policy execution. We now assess the robustness of the executed policies, by comparing the averaged return obtained over multiple policy executions with the corresponding target return. We show that the executed policies reliably obtain returns that are similar to the desired return used to condition PCN.

Results are shown in Table~\ref{tab:pcn-robustness}. The $I_\varepsilon$ indicators shows that, over all policies, the decision maker will lose at worst $0.046$ and $0.068$ normalized returns in any of the objectives for the ODE, Binomial versions respectively. On average, it will lose $0.007$, $0.024$ normalized returns respectively, which corresponds to either a total of $3633, 1059$ more hospitalizations than expected or a total of $262, 76$ less contacts than expected.

\begin{table}[t]
    \centering
    \setlength{\tabcolsep}{0.5em} 
    {\renewcommand{\arraystretch}{1.2}
    \begin{tabular}{|l|c|c|}
    \hline
        & $I_\varepsilon$ & $I_{\varepsilon-mean}$ \\
    \hline
        $\text{PCN}_\text{ARH}$  (ODE) & $0.046 \pm 0.012$ & $0.007 \pm 0.003$ \\
        \hline
        $\text{PCN}_\text{ARH}$  (Binomial) &  $0.068 \pm 0.021$ & $0.024 \pm 0.002$ \\
        \hline
        $\text{PCN}_\text{ARI}$  (ODE) & $0.056 \pm 0.013$ & $0.008 \pm 0.004$ \\
        \hline
        $\text{PCN}_\text{ARI}$  (Binomial) &  $0.058 \pm 0.012$ & $0.011 \pm 0.003$ \\
        \hline
\end{tabular}

    }
    \caption{Comparing the difference in the desired return provided to PCN and the actual return PCN obtained when executing its policy. We see that, regardless of the setting, the learned policy faithfully receives a return similar to its desired return.}
    \label{tab:pcn-robustness}
\end{table}

\section{Related work}
Reinforcement learning (RL) has been used in conjunction with epidemiological models to learn policies to limit the spread of diseases and predict the effects of possible mitigation strategies \cite{probert2019context,ernst2006clinical}. For example, RL has been used extensively in modelling and controlling the spread of influenza \cite{das2008large,libin2020,libin2018bayesian}.
 
RL and Deep RL have been used extensively as a decision making aid to reduce the spread of COVID-19. For example, to learn effective mitigation strategies \cite{ohi2020exploring}, to learn efficacy of lockdown and travel restrictions \cite{kwak2021covid} and to limit the influx of asymptomatic travellers \cite{bastani2021efficient}.

Multi-objective methods have also been deployed to learn optimal strategies to mitigate the spread of COVID-19. Wan et al. \cite{wan2021multi} implement a model-based multi-objective policy search method and demonstrate their method on COVID-19 data from China. Given the method is model-based, a model of the transition function must be learned by sampling from the environment. The method proposed by Wan et al. \cite{wan2021multi} only considers a discrete action space which limits the application of their algorithm. Wan et al. \cite{wan2021multi} use linear weights to compute a set of Pareto optimal policies. However, methods which use linear weights can only learn policies on the convex-hull of the Pareto front \cite{vamplew2008limitations}, therefore the full Pareto front cannot be learned. It is important to note the method proposed by Kompella et al. \cite{kompella2020reinforcement} considers multiple objectives. However, the objectives are combined using a weighted sum with hand-tuned weights which are determined by the authors. The weighted sum is applied by the reward function and a single objective RL method is used to learn a single optimal policy.
In contrast to previous work, our approach makes no assumptions regarding the scalarisation function of the user and is able to discover Pareto fronts of arbitrary shape.

\section{Conclusion and discussion}
Making decisions on how to mitigate epidemics has important ethical implications with respect to public health and societal burden. In this regard, it is crucial to approach this decision making from a balanced perspective, to which end we argue that multi-objective decision making is crucial. In this work, we establish a novel approach, i.e., an expert system, to study multi-faceted policies, and this approach shows great potential with respect to future epidemic control. We are aware of the ethical implications that expert systems have on the decision process and we make the disclaimer that all results based on, or derived from, the expert system that we propose should be carefully interpreted by experts in the field of public health, and in a much broader context of economics, well-being and education. We note that the work in this manuscript was conducted by a multi-disciplinary consortium that includes computer scientists and scientists with a background public health, epidemiology and bio-statistics.

In this work, we focus on the clinical outcomes of the intervention strategies and use the reduced contacts as proxy for the quality of life lost. This could be extended into more formal economic evaluations by assessing the health and monetary benefits and costs of different interventions for various stakeholders. The COVID-19 pandemic demonstrated the broad impact of infectious diseases on sectors outside health care. This stresses the need to capture a societal and thus multi-objective perspective in the decision making process on public health and health care interventions. Our learned policies confirm this, showing that focusing solely on reducing the number of hospitalizations results in taking drastic measures -- more than a thousand social interactions lost per person over the span of 4 months -- that may have a long-lasting impact on the population.

To conclude, we show that multi-objective reinforcement learning can be used to learn a wide set of high-quality policies on real-world problems, providing the decision maker with insightful and diverse alternatives, and showing the impact of taking extreme measures.

\bibliographystyle{unsrtnat}
\bibliography{references}


\appendix
\counterwithin{figure}{section}
\counterwithin{table}{section}
\counterwithin{equation}{section}

\section{Compartmental model for SARS-CoV-2}
In this work we utilise the compartmental model proposed by \citet{abrams2021modelling} and extend the model to a multi-objective Markov decision process (MOMDP). The MOMDP used in this work is outlined in the main paper. However, we describe the details of the underlying compartmental model used to model the spread of COVID-19 below. 

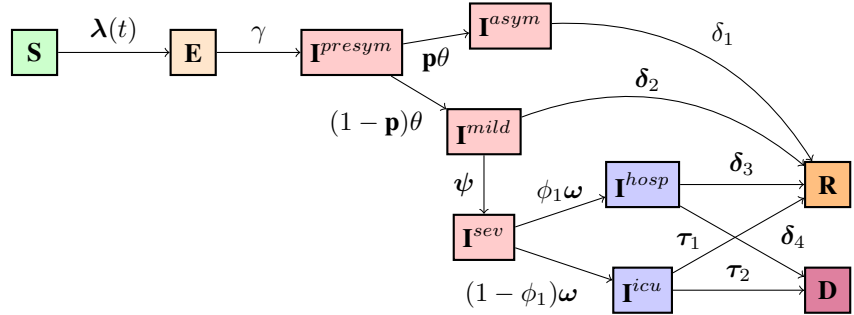
\begin{figure}[h]
    \centering
     \begin{tikzpicture}
     \draw (0.0pt, 0.0pt)node[fill=green!20, thick, minimum height=0.6cm,minimum width=0.6cm, draw](0){$\textbf{S}$};
     \draw (60.0pt, 0.0pt)node[fill=orange!20, thick, minimum height=0.6cm,minimum width=0.6cm, draw](1){$\textbf{E}$};
     \draw (120.0pt, 0.0pt)node[fill=red!20, thick, minimum height=0.6cm,minimum width=0.6cm, draw](2){$\textbf{I}^{presym}$};
     \draw (180.0pt, 10.0pt)node[fill=red!20, thick, minimum height=0.6cm,minimum width=0.6cm, draw](3){$\textbf{I}^{asym}$};
     \draw (170.0pt, -30.0pt)node[fill=red!20, thick, minimum height=0.6cm,minimum width=0.6cm, draw](4){$\textbf{I}^{mild}$};
     \draw (170.0pt, -70.0pt)node[fill=red!20, thick, minimum height=0.6cm,minimum width=0.6cm, draw](5){$\textbf{I}^{sev}$};
     \draw (230.0pt, -50.0pt)node[fill=blue!20, thick, minimum height=0.6cm,minimum width=0.6cm, draw](6){$\textbf{I}^{hosp}$};
     \draw (230.0pt, -90.0pt)node[fill=blue!20, thick, minimum height=0.6cm,minimum width=0.6cm, draw](7){$\textbf{I}^{icu}$};
     \draw (300.0pt, -50.0pt)node[fill=orange!50, thick, minimum height=0.6cm,minimum width=0.6cm, draw](8){$\textbf{R}$};
     \draw (300.0pt, -90.0pt)node[fill=purple!50, thick, minimum height=0.6cm,minimum width=0.6cm, draw](9){$\textbf{D}$};
     
     \path[->] (0) edge[] node[above]{$\boldsymbol{\lambda}(t)$} (1);
     \path[->] (1) edge[] node[above]{$\gamma$} (2);
     \path[->] (2) edge[] node[below]{$\textbf{p}\theta$} (3);
     \path[->] (2) edge[] node[near end, below left]{$(1-\textbf{p})\theta$} (4);
     \path[->] (4) edge[] node[left]{$\boldsymbol{\psi}$} (5);
     \path[->] (5) edge[] node[near end, below left]{$(1-\phi_{1})\boldsymbol{\omega}$} (7);
     \path[->] (5) edge[] node[above]{$\phi_{1}\boldsymbol{\omega}$} (6);
     \path[->] (7) edge[] node[near end, below right]{$\boldsymbol{\delta}_{4}$} (8);
     \path[->] (7) edge[] node[above]{$\boldsymbol{\tau}_{2}$} (9);
     \path[->] (6) edge[] node[above]{$\boldsymbol{\delta}_{3}$} (8);
     \path[->] (6) edge[] node[near start, below left]{$\boldsymbol{\tau}_{1}$} (9);
     \path[->] (3) edge[bend left = 30] node[above right]{$\delta_{1}$} (8);
     \path[->] (4) edge[bend left = 30] node[above left]{$\boldsymbol{\delta}_{2}$} (8);
      
    \end{tikzpicture}
    
    \caption{Schematic diagram of the compartmental model for SARS-CoV-2 presented by \citet{abrams2021modelling} which is used to derive the MOMDP.
    }
    \label{fig:ode_mdel}
\end{figure}

We utilise an adapted SEIR mathematical compartmental model proposed by Abrams et al. to contruct the MOBelCov MOMDP with deterministic and stochastic state transitions. In this model members of the population are susceptible to infection when they are in compartment $\textbf{S}$\footnote{Boldface vector notation is used to denote the multiple age groups for each compartment.}. If an individual comes into contact with an infections individual then they move to the exposed compartment, $\textbf{E}$, at a time specific rate, $\boldsymbol{\lambda} (t)$. After a period of time an exposed individual becomes infectious, where they move to the pre-symptomatic compartment, $\textbf{I}^{presym}$, at rate $\gamma$. Once infected, individuals develop mild symptoms, $\textbf{I}^{mild}$ , with probability $1 - \textbf{p}$ or do not develop any symptoms, $\textbf{I}^{asym}$, with probability $\textbf{p}$, where asymptomatic individuals recover, $\textbf{R}$, at rate $\delta_{1}$. Individuals who experience symptoms can suffer from a mild infection, $\textbf{I}^{mild}$, and recover at rate $\boldsymbol{\delta}_{2}$, or they suffer from a more severe infection, $\textbf{I}^{sev}$, at a rate $\boldsymbol{\psi}$. Individuals with a severe infection are then transferred to a hospital for treatment, $\textbf{I}^{hosp}$ with probability $\phi_{1}$. However, some individuals become critically ill and are transferred directly to the intensive care unit (ICU) with probability $1 - \psi_{1}$. Individuals in the hospital, $\textbf{I}^{hosp}$ and $\textbf{I}^{icu}$, recover at rate $\boldsymbol{\delta}_{3}$ or $\boldsymbol{\delta}_{4}$ and die at rate $\boldsymbol{\tau}_{3}$ and $\boldsymbol{\tau}_{4}$ respectively. Figure \ref{fig:ode_mdel} outlines the compartmental model defined by Abrams et al.~\cite{abrams2021modelling}.

\subsection{Deterministic Compartmental Model}
\label{sec:deterministic_comp_model}
The flows of the deterministic model are defined by a set of ordinary differential equations, which are outlined as follows:

\begin{align*}
\frac{d\textbf{S}(t)}{dt} & = -\lambda(t)(\textbf{S})(t), \\
\frac{d\textbf{E}(t)}{dt} & = \lambda(t)(\textbf{S})(t) - \gamma \textbf{E}(t), \\
\frac{d\textbf{I}^{presym}(t)}{dt} & = \gamma \textbf{E}(t) - \theta \textbf{I}^{presym}(t), \\
\frac{d\textbf{I}^{asym}(t)}{dt} & = \theta p \textbf{I}^{presym}(t) - \delta_{1} \textbf{I}^{asym}(t), \\
\frac{d\textbf{I}^{mild}(t)}{dt} & = \theta (1 - p) \textbf{I}^{presym}(t) - \{ \psi + \omega_{2} \} \textbf{I}^{mild}(t), \\
\frac{d\textbf{I}^{sev}(t)}{dt} & = \psi \textbf{I}^{mild}(t) - \omega \textbf{I}^{sev}(t), \\
\frac{d\textbf{I}^{hosp}(t)}{dt} & = \phi_{1} \omega \textbf{I}^{sev}(t) - \{ \delta_{3} + \tau_{1} \} \textbf{I}^{hosp}(t), \\
\frac{d\textbf{I}^{icu}(t)}{dt} & = (1 - \phi_1) \omega \textbf{I}^{sev}(t) - \{ \delta_{4} + \tau_{2} \} \textbf{I}^{icu}(t), \\
\frac{d\textbf{D}(t)}{dt} & = \tau_{1} \textbf{I}^{hosp}(t) - \tau_{2} \textbf{I}^{icu}(t), \\
\frac{d\textbf{R}(t)}{dt} & = \delta_{1} \textbf{I}^{asym}(t) + \delta_{2} \textbf{I}^{mild}(t) + \delta_{3} \textbf{I}^{hosp}(t) + \delta_{4} \textbf{I}^{icu}(t)
\end{align*}

where, for example, $\textbf{S} = (S_{1}(t), S_{2}(t), ..., S_{k}(t))^{T}$ is the vector representing the susceptible members of the population of each age group $k$ at time $t$. 

In this set of ordinary differential equations, each state variable represents a vector over all age groups for a particular compartment at time $t$.
Infection dynamics are governed by an age-specific force of infection $\foi$:
\begin{equation}
    \foi(k,t) = \sum_{k'=1}^{K}\beta(k,k')I_{k'}(t),
\end{equation}
where $k$ is the respective age group of a total of K age groups, and $\beta(k,k')$ is the time-invariant transmission rate that encodes the average per capita rate at which an infectious individual in age group $k$ makes an effective contact with a susceptible individual in age group $k'$, per unit of time. 

Under the social contact hypothesis \cite{wallinga2006using}, we have that:
\begin{equation}
\beta(k,k') = q \cdot C(k,k'),
\end{equation}
where q is a proportionality factor and the matrix $C$ denotes the social mixing behaviour within and between different age groups in the population, and is referred to as a social contact matrix. 
Following \citet{abrams2021modelling}, we rely on distinct social contact matrices for symptomatic and asymptomatic individuals, respectively $C_{s}$ and $C_{a}$. Therefore it is possible to define the transmission rates for both symptomatic and asymptomatic individuals as follows:

\begin{equation}
\boldsymbol{\beta}_{s}(k, k') = q_{s} \cdot C_{s}(k, k'),    
\end{equation}
and
\begin{equation}
\boldsymbol{\beta}_{a}(k, k') = q_{a} \cdot C_{a}(k, k').  
\end{equation}
Moreover, the age-dependent force of infection can be defined as follows:
\begin{equation}
    \boldsymbol{\lambda}(t) = \boldsymbol{\beta}_{a} \times \{ \textbf{I}^{presym}(t) + \textbf{I}^{asym}(t) \} +
    \boldsymbol{\beta}_{s} \times \{ \textbf{I}^{mild}(t) + \textbf{I}^{sev}(t) \},
\end{equation}
where $\boldsymbol{\lambda}(t) = (\lambda(1, t), \lambda(1, t), ..., \lambda(K, t))$. For all further information about the different compartments and parameters please refer to the work of \citet{abrams2021modelling}

To create a version of MOBelCov with a deterministic transition function, $M$, we focus on the deterministic model proposed by \citet{abrams2021modelling}. Given that the transitions within the compartmental model are deterministic it is possible to utilise the highlighted model transitions for MOBelCov. Applying the contact matrix $\hat{C}$ to the model state $s_{m}$ progresses the model for one timestep and returns a new compartmental model state $s'_{m}$. Given $s_{m}$ and $\hat{C}$, the  deterministic compartmental model returns $s'_{m}$ with probability of $1$. Therefore, it is possible to use this process as a deterministic transition function, $M$, for MOBelCov. 

\subsection{Stochastic Compartmental Model}
Intervening in the spread of the virus by, for example, reducing social contacts or government interventions introduces uncertainty in the further course of the outbreak. Therefore, to understand how this uncertainty affect the spread of the disease we introduce a stochastic component model which can model the uncertainty generated by interventions in social contacts. 

By formulating the set of differential equations defined above, as a chain-binomial, we can obtain stochastic trajectories from this model \cite{bailey1975mathematical}. A chain-binomal model assumes a stochastic model where infected individuals are generated by some underlying probability distribution. For the stochastic model we consider a time interval $(t, t +h]$, where $h$ is defined as the length between two consecutive time points. Similar to \citet{abrams2021modelling}, in this work we set $h = \frac{1}{24}$. \citet{abrams2021modelling} define the set of differential equations as a chain binomial as follows:

\begin{align*}
\textbf{S}_{t+h} (k) & = \textbf{S}_{t} (k) - \textbf{E}_{new, t+h}(k),\\
\textbf{E}_{t+h}(k) & = \textbf{E}_{t}(k) + \textbf{E}_{new, t+h}(k) - \textbf{I}^{presym}_{new, t + h} (k),\\
\textbf{I}^{presym}_{t+h}(k) & = \textbf{I}^{presym}_{t}(k) + \textbf{I}^{presym}_{new, t + h}(k) - \textbf{I}^{asym}_{new, t + h}(k) - \textbf{I}^{mild}_{new, t + h}(k),\\
\textbf{I}^{asym}_{t+h}(k) & = \textbf{I}^{asym}_{t}(k) + \textbf{I}^{asym}_{new, t + h}(k) - \textbf{R}^{asym}_{new, t + h}(k),\\
\textbf{I}^{mild}_{t+h}(k) & = \textbf{I}^{mild}_{t}(k) + \textbf{I}^{mild}_{new, t + h}(k) - \textbf{I}^{sev}_{new, t + h}(k) - \textbf{R}^{mild}_{new, t + h}(k),\\
\textbf{I}^{sev}_{t+h}(k) & = \textbf{I}^{sev}_{t}(k) + \textbf{I}^{sev}_{new, t + h}(k) - \textbf{I}^{hosp}_{new, t + h}(k) - \textbf{I}^{icu}_{new, t + h}(k),\\
\textbf{I}^{hosp}_{t+h}(k) & = \textbf{I}^{hosp}_{t}(k) + \textbf{I}^{hosp}_{new, t + h}(k) - \textbf{D}^{hosp}_{new, t + h}(k) - \textbf{R}^{hosp}_{new, t + h}(k),\\
\textbf{I}^{icu}_{t+h}(k) & = \textbf{I}^{icu}_{t}(k) + \textbf{I}^{icu}_{new, t + h}(k) - \textbf{D}^{icu}_{new, t + h}(k) - \textbf{R}^{icu}_{new, t + h}(k),\\
\textbf{D}_{t+h}(k) & = \textbf{D}_{t}(k) + \textbf{D}^{hosp}_{new, t + h}(k) + \textbf{D}^{icu}_{new, t + h}(k),\\
\textbf{R}_{t+h}(k) & = \textbf{R}_{t}(k) + \textbf{R}^{asym}_{new, t + h}(k) + \textbf{R}^{mild}_{new, t + h}(k) + \textbf{R}^{hosp}_{new, t + h}(k) + \textbf{R}^{icu}_{new, t + h}(k)
\end{align*}

where,

\begin{align*}
\textbf{E}_{new, t+h} & \sim \textit{Binomial} \left( \textbf{S}_{t} (k), p^{*}_t (k) = 1 - \{ 1 -  p^{*}_t (k)\}^{\textbf{I}_{t}} \right), \\
p^{*}_t (k) & = 1 - exp \left[ -h \sum^{K}_{k' = 1} \beta_{asym}(k, k')\{ \textbf{I}^{asym}_{t}(k')\} + \beta_{sym}(k, k') \{\textbf{I}^{mild}_{t}(k')+\textbf{I}^{sev}_{t}(k')\} \right],\\
\textbf{I}^{presym}_{new, t+h} (k)&  \sim Binomial\left( \textbf{I}^{presym}_{t}(k), 1 - exp(-hp(k)\theta) \right),\\
\textbf{I}^{mild}_{new, t+h} (k) & \sim Binomial\left( \textbf{I}^{presym}_{t}(k), 1 - exp\left[-h \{1-p(k)\}\theta \right]) \right),\\
\textbf{I}^{sev}_{new, t+h} (k) & \sim Binomial\left( \textbf{I}^{mild}_{t}(k), 1 - exp\{-h\psi(k)\}) \right),\\
\textbf{I}^{hosp}_{new, t+h} (k) & \sim Binomial\left( \textbf{I}^{sev}_{t}(k), 1 - exp\{-h\phi_{1}(k)\omega(k)\}) \right),\\
\textbf{I}^{icu}_{new, t+h} (k) & \sim Binomial\left( \textbf{I}^{sev}_{t}(k), 1 - exp\left[-h \{1-\phi_{1}(k)\}\omega(k) \right] \right),\\
\textbf{D}^{hosp}_{new, t+h} (k) & \sim Binomial\left( \textbf{I}^{hosp}_{t}(k), 1 - exp\{-h\tau_{1}(k) \} \right),\\
\textbf{D}^{icu}_{new, t+h} (k) & \sim Binomial\left( \textbf{I}^{icu}_{t}(k), 1 - exp\{-h\tau_{2}(k) \} \right),\\
\textbf{R}^{asym}_{new, t+h} (k) & \sim Binomial\left( \textbf{I}^{asym}_{t}(k), 1 - exp\left(-h\delta_{2}(k) \right) \right), \\
\textbf{R}^{hosp}_{new, t+h} (k) & \sim Binomial\left( \textbf{I}^{hosp}_{t}(k), 1 - exp\{-h\delta_{3}(k) \} \right),\\
\textbf{R}^{icu}_{new, t+h} (k) & \sim Binomial\left( \textbf{I}^{icu}_{t}(k), 1 - exp\{-h\delta_{4}(k) \} \right).
\end{align*}

Given MOBelCov also calculates new hospitalisations, $\textbf{H}^{new}$, we define $\textbf{H}^{new}$ for the stochastic compartmental model as follows:
\[
\textbf{H}^{new}_{t+h}(k) = \textbf{H}^{new}_{t}(k) + \textbf{I}^{hosp}_{new, t+h} (k).
\]
For more details on this model and the chain-binomial representation of the differential equations, we refer the reader to the work of \citet{abrams2021modelling}.

To create a version of MOBelCov with a stochastic transition function, $M$, we utilise the stochastic compartmental model outlined above. Given the transitions within the compartmental model are derived by an underlying probability distribution it is possible to utilise the stochastic compartmental model transitions for MOBelCov. As previously outlined in Section \ref{sec:deterministic_comp_model} the contact matrix $\hat{C}$ applied the model state $s_{m}$ progresses the model and returns a new model state $s'_{m}$. Given the underlying model dynamics are governed in a probabilistic manner, the model returns $s'_{m}$ stochastically. Therefore, it is possible to use this process as a stochastic transition function, $M$, for MOBelCov. 

\subsection{A Note on Model Parameters}

The model is parameterised using the mean of the posteriors as reported by \citet{abrams2021modelling}.

The population size for each of the considered age groups was taken from the Belgian statistical agency STATBEL\footnote{\url{https://statbel.fgov.be/nl/themas/bevolking/structuur-van-de-bevolking\#figures}}. To initialise the model, we used the number of confirmed cases until 13 March 2020 \cite{abrams2021modelling}, as reported by the Belgian agency for public health Sciensano\footnote{\url{https://epistat.wiv-isp.be/covid/}}.

\subsection{Modelling interventions}
\label{sec:interventions}
In order to model different types of interventions, we follow \citet{abrams2021modelling}. Firstly, 
to consider distinct exit scenarios, we alter the social contact matrices to reflect a contact reduction in a particular age group. Secondly, we assume that compliance to the interventions is gradual and model this using a logistic compliance function. We use the logistic compliance function in function of time $t$ proposed by Abrams et al.,
\begin{equation}
c(t,t_I) = \frac{\exp(\beta^*_0 + \beta^*_1(t-t_I))}{1+\exp(\beta^*_0 + \beta^*_1(t-t_I))},
\end{equation}
where $t_I$ indicates the time the intervention started \cite{abrams2021modelling}.
We initialise $\beta^*_1$ to the value estimated in by Abrams et al. and choose $\beta^*_0=-5$, as an intercept to have $c(t)=0$ for $t=0$, in correspondence with Figure~F2 in the Supplementary Information of \citet{abrams2021modelling}.


\section{Additional results}
\subsection{Learned coverage sets}

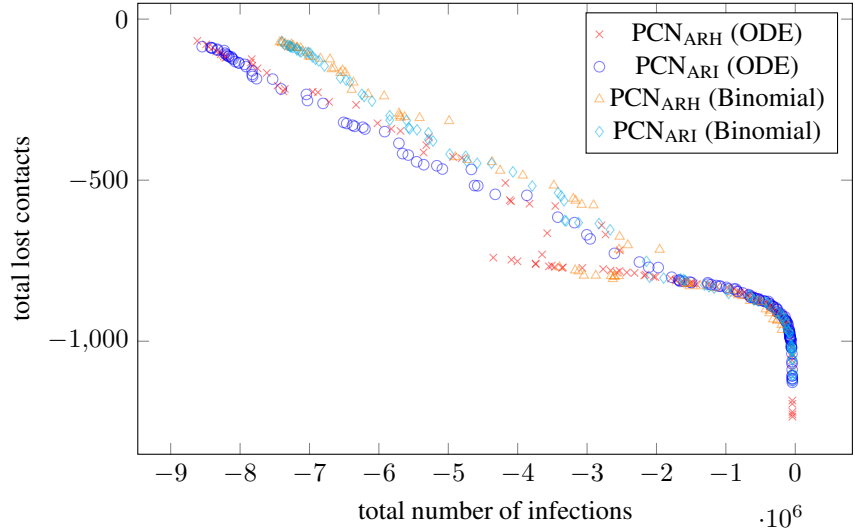
\begin{figure}[h]
    \centering
    \begin{tikzpicture}
    \pgfplotsset{
        scale only axis,
        height=6cm,
        xlabel={total number of daily new hospitalizations},
        ylabel={total lost contacts}
    }
    \begin{axis}[
    legend entries={{$\text{PCN}_\text{ARH}$ (ODE)}, {$\text{PCN}_\text{ARI}$ (ODE)}, {$\text{PCN}_\text{ARH}$ (Binomial)}, {$\text{PCN}_\text{ARI}$ (Binomial)}, {Fixed}},
    width=9.5cm
    ]
    \addplot[only marks,draw opacity=0.5,mark=x,red] table [x=o_1, y=o_5, col sep=comma] {data/cs/arh/cs_ode.csv};
    \addplot[only marks,draw opacity=0.5,mark=o,blue] table [x=o_1, y=o_5, col sep=comma] {data/cs/ari/cs_ode.csv};
    \addplot[only marks,draw opacity=0.5,mark=triangle,orange] table [x=o_1, y=o_5, col sep=comma] {data/cs/arh/cs_binomial.csv};
    \addplot[only marks,draw opacity=0.5,mark=diamond,cyan] table [x=o_1, y=o_5, col sep=comma] {data/cs/ari/cs_binomial.csv};
    \addplot[only marks,mark=+] table [x=o_0, y=o_1, col sep=comma] {data/cs/cs_fixed.csv};
    \end{axis}
\end{tikzpicture}
    \caption{The Pareto front of policies discovered by PCN, showing the different compromises between the number of hospitalizations and the number of lost contacts. Although $\text{PCN}_\text{ARI}$ was trained on the number of infections, it still shows a competitive coverage set w.r.t. hospitalizations.}
    \label{fig:pareto-front-arh}
\end{figure}

In Figure~\ref{fig:pareto-front-arh}, we display the coverage set with respect to the number of hospitalizations of the learned policies for both the ODE and Binomial variants, as the Binomial variants were omitted in the main paper for clarity. When trained on $\mathbf{R}_{\text{ARH}}$, the Binomial coverage set is on par with its ODE counterpart. On the other hand, the coverage sets when trained on $\mathbf{R}_{\text{ARI}}$ are slightly worse, as confirmed by the evaluation metrics in Table~\ref{tab:coverage-set}.

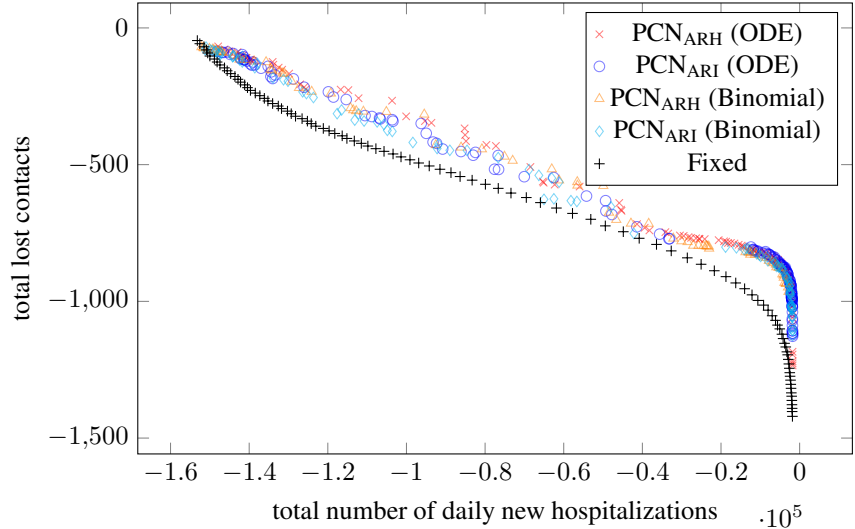
\begin{figure}[h]
    \centering
    \begin{tikzpicture}
    \pgfplotsset{
      scale only axis,
      height=6cm,
      xlabel={total number of infections},
      ylabel={total lost contacts}
    }
  \begin{axis}[
    legend entries={{$\text{PCN}_\text{ARH}$ (ODE)}, {$\text{PCN}_\text{ARI}$ (ODE)}, {$\text{PCN}_\text{ARH}$ (Binomial)}, {$\text{PCN}_\text{ARI}$ (Binomial)}},
    width=9.5cm
  ]
  \addplot[only marks,draw opacity=0.5,mark=x,red] table [x=o_0, y=o_5, col sep=comma] {data/cs/arh/cs_ode.csv};
  \addplot[only marks,draw opacity=0.5,mark=o,blue] table [x=o_0, y=o_5, col sep=comma] {data/cs/ari/cs_ode.csv};
  \addplot[only marks,draw opacity=0.5,mark=triangle,orange] table [x=o_0, y=o_5, col sep=comma] {data/cs/arh/cs_binomial.csv};
  \addplot[only marks,draw opacity=0.5,mark=diamond,cyan] table [x=o_0, y=o_5, col sep=comma] {data/cs/ari/cs_binomial.csv};
  \end{axis}
\end{tikzpicture}
    \caption{The Pareto front of policies discovered by PCN, showing the different compromises between the number of hospitalizations and the number of lost contacts. Although $\text{PCN}_\text{ARI}$ was trained on the number of infections, it still shows a competitive coverage set w.r.t. hospitalizations.}
    \label{fig:pareto-front-ari}
\end{figure}

Moreover, we also show the coverage sets with respect to the number of infections in Figure~\ref{fig:pareto-front-ari}. Interestingly, both variants trained on the Binomial model learn a coverage set that almost completely dominates the variants trained on the ODE model. This is due to the stochasticity of the Binomial model. PCN stores trajectories in its dataset, and keep a selection for training. Due to the stochasticity in the transition function, some trajectories might result in higher returns, even though the same policy was applied. Due to the high variability in infections, this difference might become significant. This is also why the variants using the Binomial model show a worse $I_{\varepsilon}$ and $I_{\varepsilon}-mean$ than their ODE counterpart: their desired return stems from a single trajectory, while we evaluate the policies over multiple trajectories. The resulting average return is then different than the desired return. Section~\ref{sec:arh-vs-ari} explains why some policies trained on $\mathbf{R}_{\text{ARH}}$ show suboptimal behavior with respect to the number of infections.

\subsection{Policy executions}

PCN learns a coverage set containing more than 100 different policies. To gain a better insight about their behavior, and how they differ from each other, we plot executions of each learned policy in Figure~\ref{fig:policy-executions-ode}. The plots are displayed from the least restrictive policy in terms of social burden to the most restrictive one.

We notice that policies~60 to 140 display similar behavior. At first, they all completely restrict social contact, effectively continuing the lockdown. Afterwards, they progressively lessen these restrictions. What differs in these policies is the timing and the speed at which the restrictions are lessened. These policies correspond to the right-most part of the coverage set displayed in Figure~\ref{fig:pareto-front-arh}, which shows many alternative for the $[0, 40000]$ hospitalizations segment.

\begin{figure}
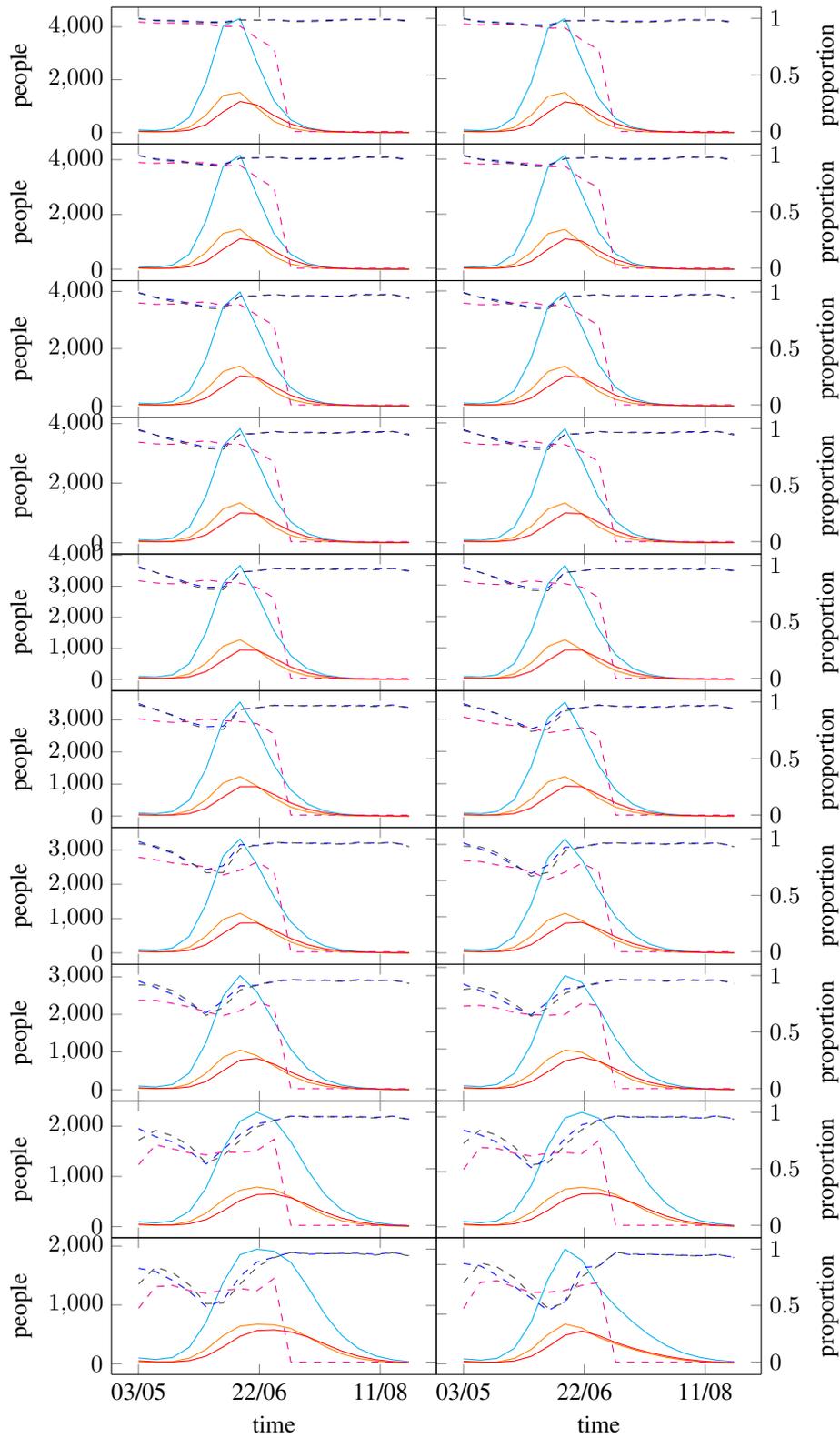

    \centering
    \plotPolicies{0}{19}
    \caption{Execution of policies 1 to 20.}
    \label{fig:policy-executions-ode}
\end{figure}
\begin{figure}
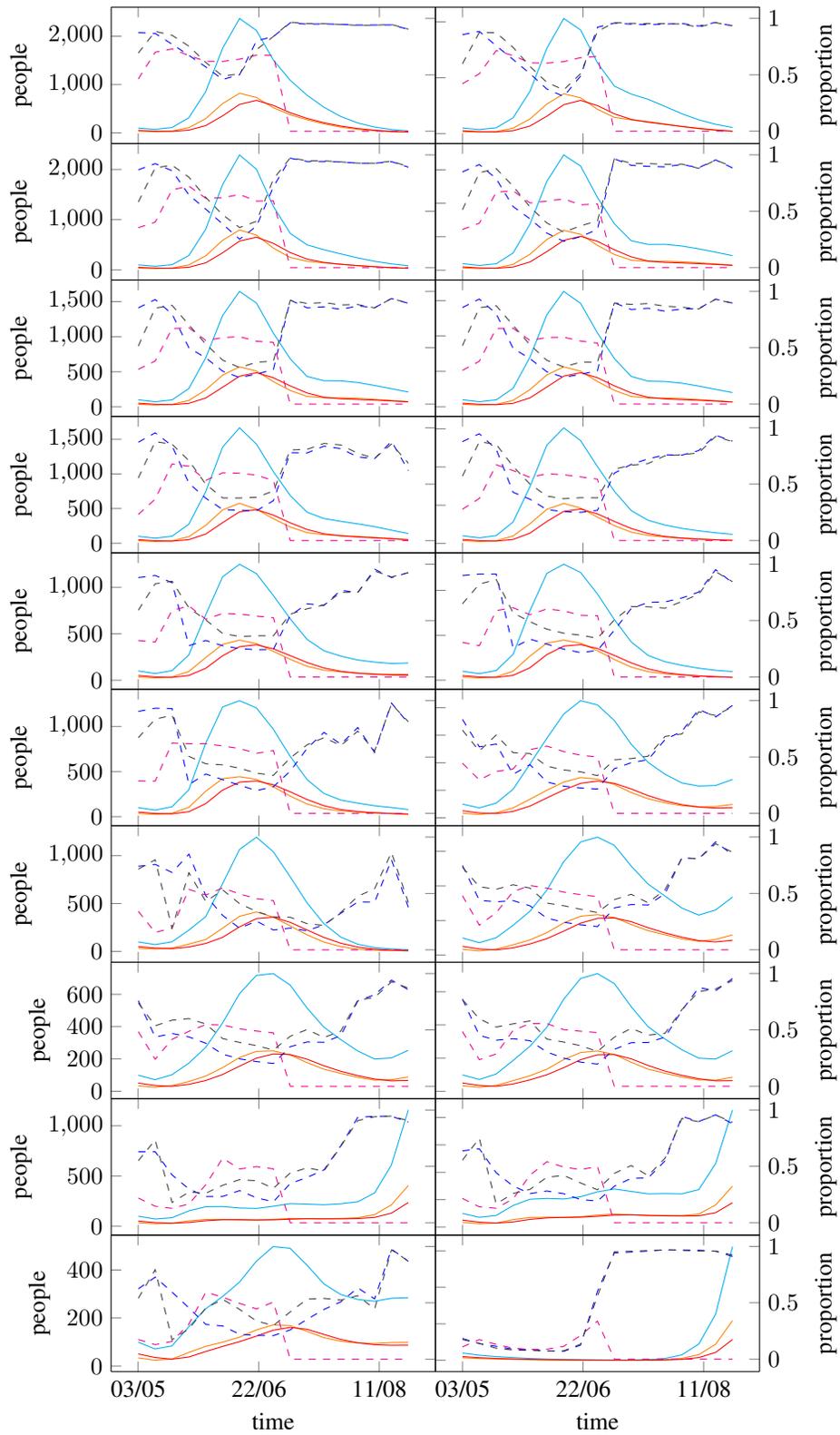
\ContinuedFloat
    \centering
    \plotPolicies{20}{39}
    \caption{Execution of policies 21 to 40.}
    \label{fig:policy-executions-ode}
\end{figure}
\begin{figure}
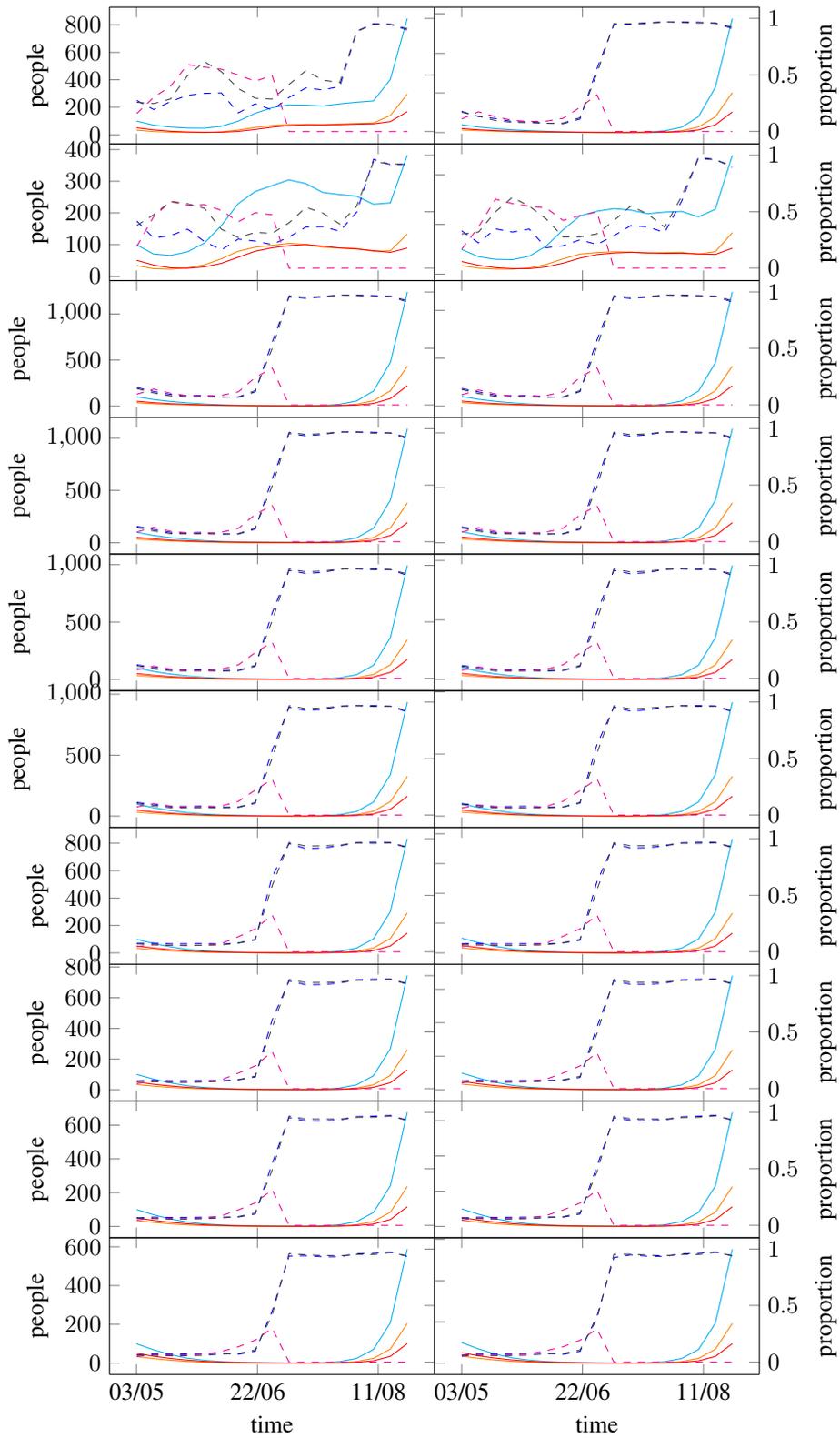
\ContinuedFloat
    \centering
    \plotPolicies{40}{59}
    \caption{Execution of policies 41 to 60.}
    \label{fig:policy-executions-ode}
\end{figure}
\begin{figure}
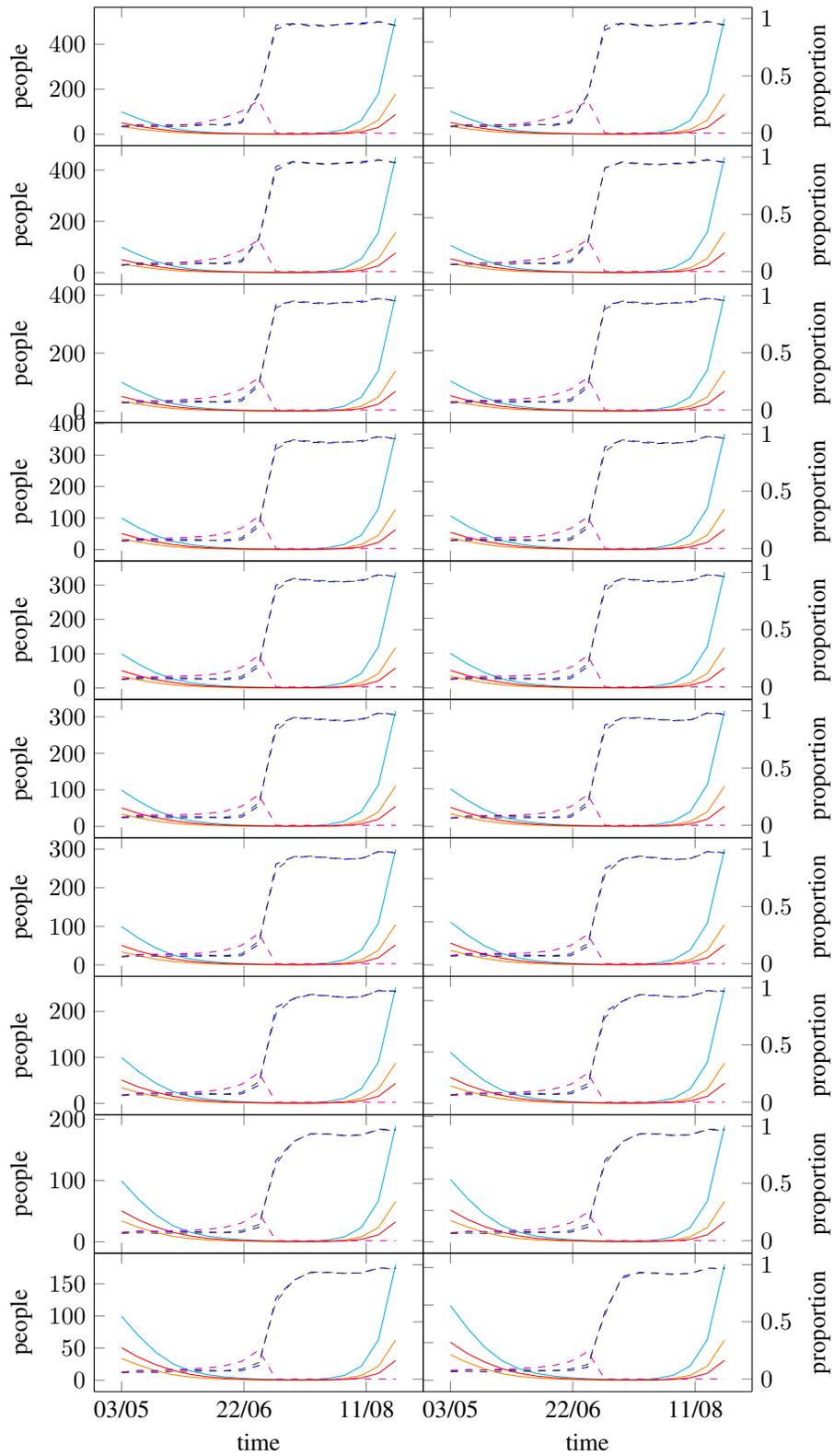
\ContinuedFloat
    \centering
    \plotPolicies{60}{79}
    \caption{Execution of policies 61 to 80.}
    \label{fig:policy-executions-ode}
\end{figure}
\begin{figure}
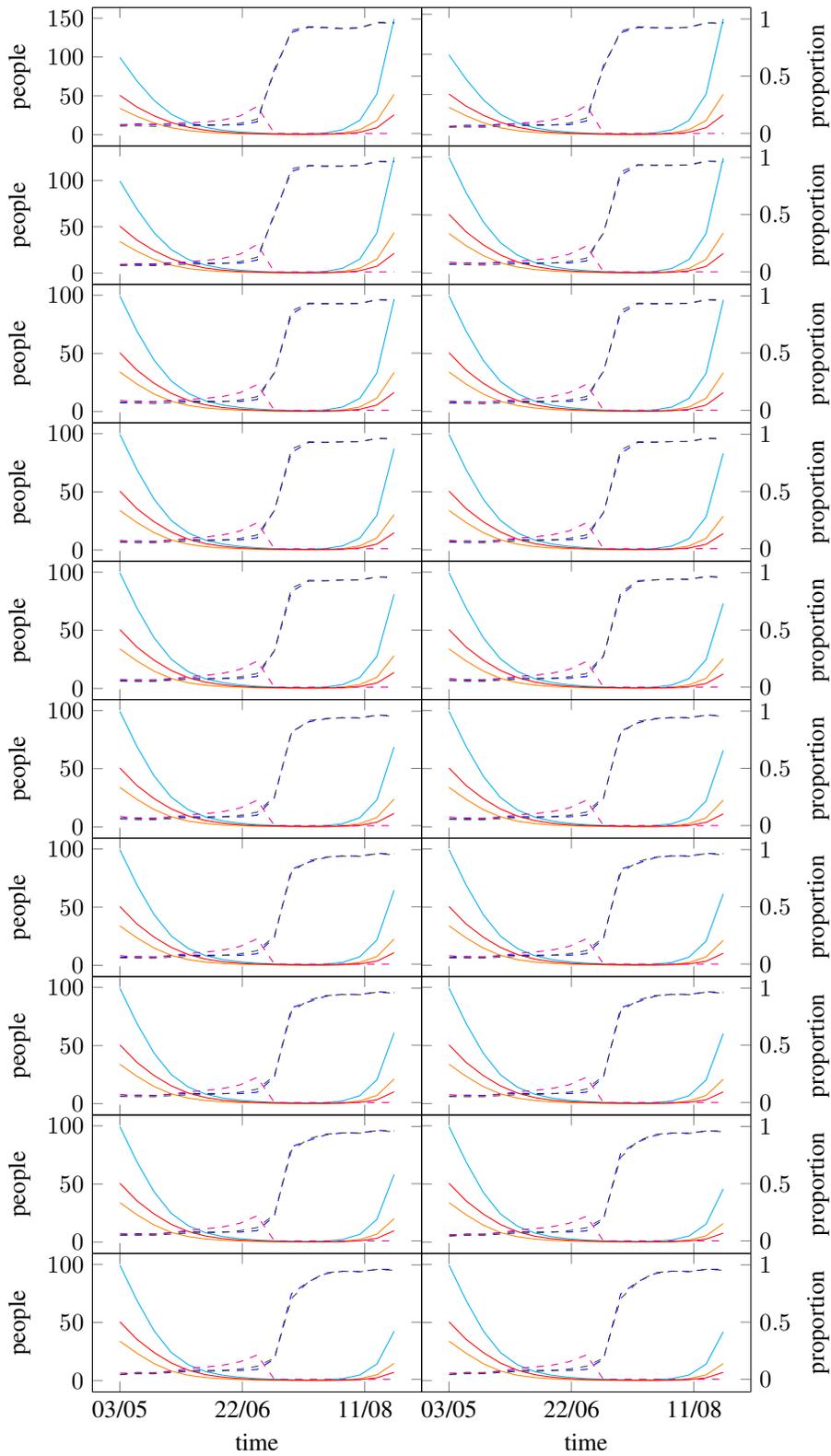
\ContinuedFloat
    \centering
    \plotPolicies{80}{99}
    \caption{Execution of policies 81 to 100.}
    \label{fig:policy-executions-ode}
\end{figure}
\begin{figure}
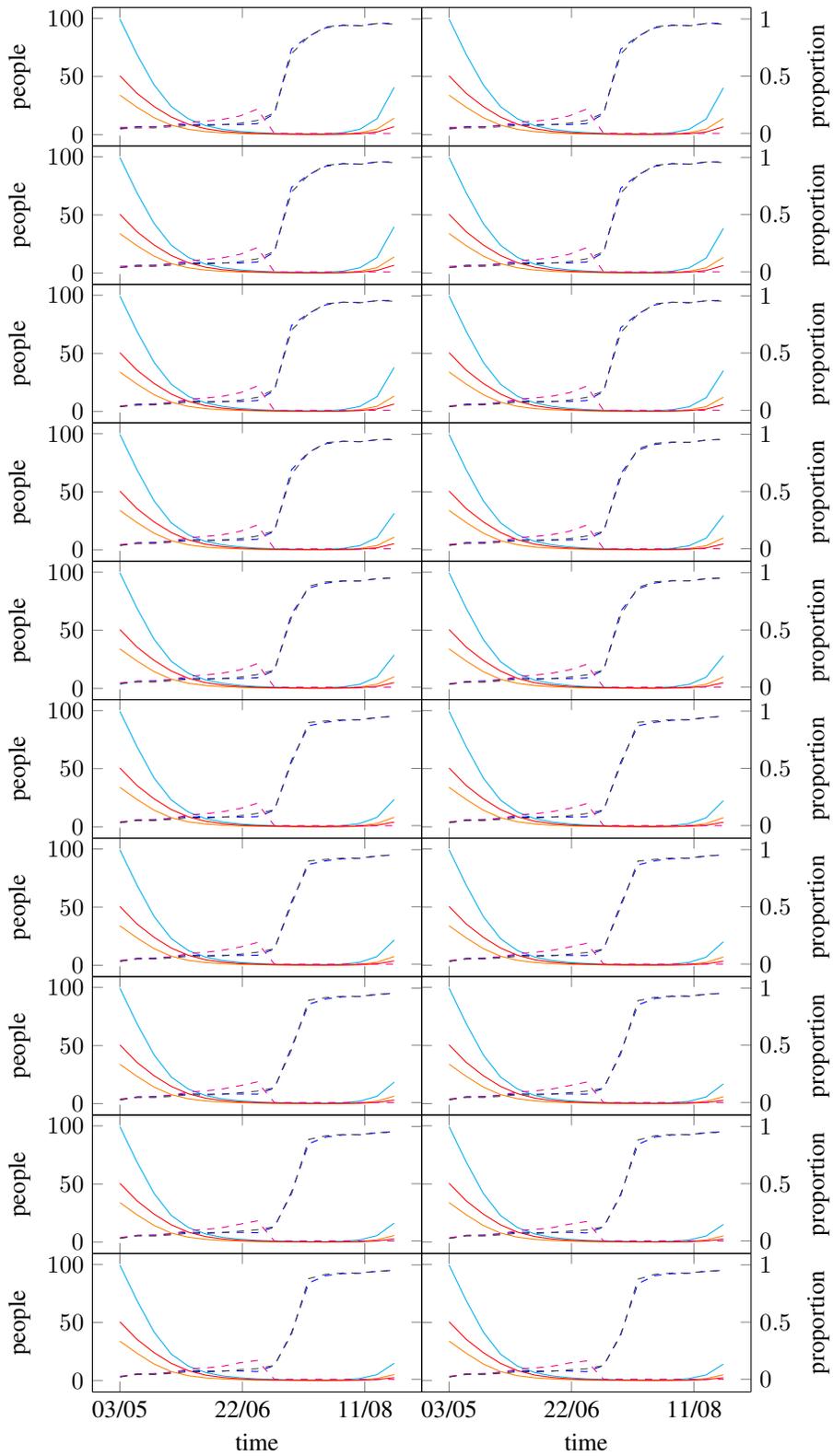
\ContinuedFloat
    \centering
    \plotPolicies{100}{119}
    \caption{Execution of policies 101 to 120.}
    \label{fig:policy-executions-ode}
\end{figure}
\begin{figure}
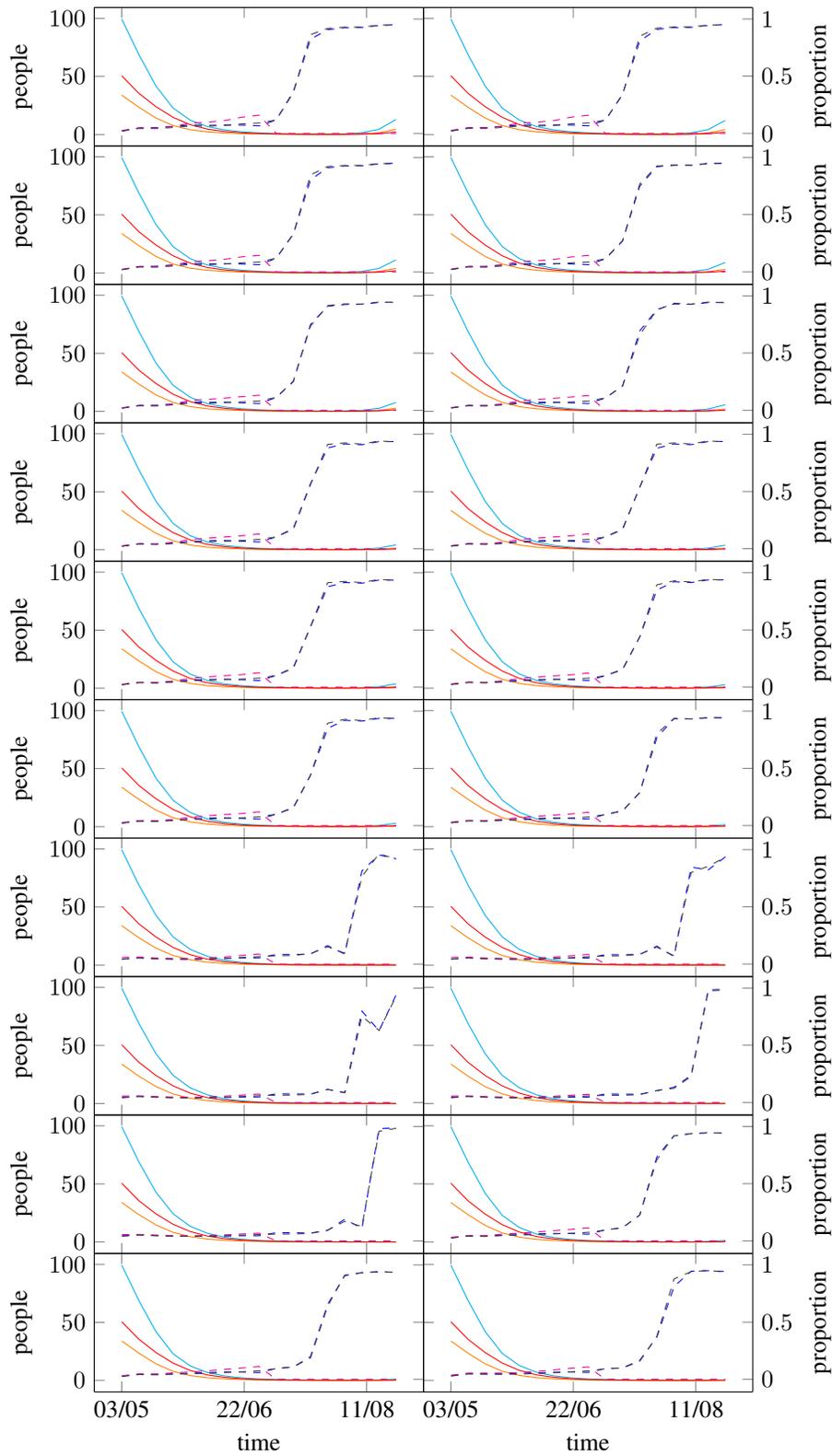
\ContinuedFloat
    \centering
    \plotPolicies{120}{139}
    \caption{Execution of policies 121 to 140.}
    \label{fig:policy-executions-ode}
\end{figure}

\subsection{Experiment parameters}

We used the same hyper-parameters across all experiments. Each experiment resulted in 5 independent trials. Finally, we performed a grid-search over possible hyper-parameter values. All hyper-parameters used and their possible values explored during grid-search are displayed in Table~\ref{tab:pcn-hyperparameters}.

\begin{table}[t]
    \centering
    \begin{tabular}{|c|c|c|}
        \hline
        hyper-parameter & value & grid-search \\
        \hline
        learning rate & $0.001$ &  \\
        total training timesteps & $300000$  & \\
        batch size & $256$ & $256, 1024$ \\
        model updates & $50$ & \\
        episodes between updates & $10$ & \\
        ER size (in episodes) & $1000$ & $400, 500, 1000$ \\
        initial random episodes & $200$ & $50, 200$ \\
        exploration noise & $0.1$ & $0, 0.1, 0.2$ \\
        desired return noise & $0.2$ & $0, 0.1, 0.2$ \\
        reward scaling & $[10000, 100]$ & \\
        \hline
    \end{tabular}
    \caption{The different hyper-parameters used by our extension of PCN. The right-most column also shows, when applicable, the different values tried during grid-search.}
    \label{tab:pcn-hyperparameters}
\end{table}

\begin{table}[h]
    \centering
    \begin{tabular}{|c|c|c|c|c|c|c|}
    \hline
        variant & $sc_{emb}$      & $sm_{emb}$   & $sh_{emb}$   & $s_{emb}$    & $c_{emb}$    & $fc$ \\
    \hline
                & conv1d(10,20)   & linear(3,64) & linear(1,64) & linear(64,64)& linear(3,64) & linear(64,64)\\
                & relu            & sigmoid      & sigmoid      & sigmoid      & sigmoid      & relu\\
    conv1d-small  & conv1d(20,20)   &              &              &              &              & linear(64,3)\\
                & relu            &              &              &         &         &      \\
                & linear(100,64)  &              &              &         &         &      \\
                & sigmoid         &              &              &         &         &      \\
    \hline
                & conv1d(10,20)   & linear(3,64) & linear(1,64) & linear(64,64)& linear(3,64) & linear(64,64)\\
                & relu            & relu         & relu         & relu         & sigmoid      & relu\\
    conv1d-big & conv1d(20,20)   & linear(64,64)& linear(64,64)&              &              & linear(64,3)\\
                & relu            & sigmoid      & sigmoid      &              &              &      \\
                & linear(100,64)  &          &          &         &         &      \\
                & sigmoid         &          &          &         &         &      \\
    \hline
                & linear(130,64)  & linear(3,64) & linear(1,64) & linear(64,64)& linear(3,64) & linear(64,64)\\
    dense-small & sigmoid         & sigmoid      & sigmoid      & sigmoid      & sigmoid      & relu\\
                &                 &              &              &              &              & linear(64,3)\\
    \hline
                & linear(130,64)  & linear(3,64) & linear(1,64) & linear(64,64)& linear(3,64) & linear(64,64)\\
    dense-big & relu            & relu         & relu         & relu         & sigmoid      & relu\\
                & linear(64,64)   & linear(64,64)& linear(64,64)&              &              & linear(64,3)\\
                & sigmoid         & sigmoid      & sigmoid      &              &              &      \\
    \hline
\end{tabular}

    \caption{The 4 different neural network architectures explored for our experiments. All the displayed results use the \texttt{conv1d-big} variant.}
    \label{tab:nn-architecture}
\end{table}

\subsection{Neural network architecture}

Next to the hyper-parameter search, we also performed a grid search over 4 different neural network architectures. All the architectures have the same structure. We use a compartment embedding $sc_{emb}$, a SCM embedding $sm_{emb}$ and a school-holidays embedding $sh_{emb}$ that take as inputs the compartment, the previous $p_w, p_s, p_l$ values (as they fully define the SCM $\hat{C}$) and a boolean flag for school holidays, respectively. All these embeddings have a same-sized embedding of 64, which are multiplied together to form the full state embedding. This state-embedding is used as input for another network, $s_{emb}$. Additionally, we use a common embedding $c_{emb}$ for the concatenation of the desired return and horizon. Finally, the results of $s_{emb}$ and $c_{emb}$ are multiplied together, before passing through a fully connected network $fc$ that has 3 outputs, one for $p_w, p_s, p_l$ respectively.

All the architectures of the different components are displayed in Table~\ref{tab:nn-architecture}. The variant used in all experiments is \texttt{conv1d-big}.

\section{Related Work}
RL has also been used to learn effective mitigation strategies which limit the spread of COVID-19 \cite{ohi2020exploring}. Kwak et al. \cite{kwak2021covid} use deep RL, with an SIRD model, to learn efficacy of lockdown and travel restrictions in controlling the COVID-19 pandemic using data collected from various government organisations.  Bastani et al. \cite{bastani2021efficient} use a RL algorithm called Eva to limit the influx of asymptomatic travellers infected with SARS-CoV-2 to Greece. Kompella et al. \cite{kompella2020reinforcement} define a COVID-19 simulator and a methodology to learn mitigation policies that minimize the economic impact without overwhelming the hospital capacity. RL has also been used to learn effective COVID-19 vaccination distribution strategies \cite{beigi2021application,awasthi2020vacsim} and to create decision support systems for mitigating the spread of COVID-19.

\end{document}